\documentclass[10pt,twocolumn,letterpaper]{article}

\usepackage[pagenumbers]{cvpr} 
\makeatletter
\@namedef{ver@everyshi.sty}{}
\makeatother

\usepackage[dvipsnames,svgnames,x11names]{xcolor}
\usepackage{tikz}
\usetikzlibrary{arrows.meta,shapes,calc,matrix,fit,positioning,backgrounds,decorations.markings}
\usepackage{pgfplots}
\usepackage{pgfplotstable}
\pgfplotsset{compat=1.9}
\usepackage{xstring}

\usepgfplotslibrary{external}

\IfBeginWith*{\jobname}{fig/extern/}{\finalcopy}{}

\tikzstyle{every picture}+=[
	remember picture,
	every text node part/.style={align=center},
	every matrix/.append style={ampersand replacement=\&},
]
\tikzstyle{tight} = [inner sep=0pt,outer sep=0pt]
\tikzstyle{node}  = [draw,circle,tight,minimum size=12pt,anchor=center]
\tikzstyle{op}    = [draw,circle,tight]
\tikzstyle{dot}   = [fill,draw,circle,inner sep=1pt,outer sep=0]
\tikzstyle{pt}    = [fill,draw,circle,inner sep=1.5pt,outer sep=.2pt]
\tikzstyle{box}   = [draw,thick,rectangle,inner sep=3pt]
\tikzstyle{high}  = [black!60]
\tikzstyle{group} = [high,box,opacity=.5]
\tikzstyle{rectc} = [tight,transform shape]
\tikzstyle{rect}  = [rectc,anchor=south west]

\tikzset{every mark/.append style={solid}}
\pgfplotsset{
	grid=both, width=\columnwidth, try min ticks=5,
	every axis/.append style={font=\small},
	every axis plot/.append style={thick,mark=none,mark size=1.8,tension=0.18},
	legend cell align=left, legend style={fill opacity=0.8},
	xticklabel={\pgfmathprintnumber[assume math mode=true]{\tick}},
	yticklabel={\pgfmathprintnumber[assume math mode=true]{\tick}},
	nodes near coords math/.style={
		nodes near coords={\pgfmathprintnumber[assume math mode=true]{\pgfplotspointmeta}},
	},
}

\pgfplotsset{
	dash/.style={mark=o,dashed,opacity=0.6},
	dott/.style={mark=o,dotted,opacity=0.6},
	nolim/.style={enlargelimits=false},
	plain/.style={every axis plot/.append style={},nolim,grid=none},
}

\usepackage{float}
\usepackage{graphicx}
\usepackage{amsmath}
\usepackage{amssymb}
\usepackage[utf8]{inputenc}
\usepackage{mathtools}
\usepackage[dvipsnames]{xcolor}
\usepackage{color, colortbl}
\usepackage{transparent}
\usepackage{enumitem}
\usepackage{xspace}
\usepackage{bm}
\usepackage{array,booktabs}
\usepackage{makecell}
\usepackage{multirow}
\usepackage[numbers,sort&compress]{natbib}
\usepackage[normalem]{ulem}
\usepackage{verbatim}
\usepackage{kantlipsum}
\usepackage{wrapfig}
\usepackage{dsfont}
\usepackage{comment}
\usepackage{cuted}
\usepackage{pifont}
\usepackage{capt-of}
\usepackage[skip=2pt]{parskip}
\usepackage[accsupp]{axessibility}
\usepackage[pagebackref,breaklinks,colorlinks,linkcolor=blue]{hyperref}
\usepackage[ruled,vlined,linesnumbered]{algorithm2e}

\usepackage[pagebackref,breaklinks,colorlinks]{hyperref}

\newcommand{\methodname}{Skip-Attention}
\newcommand{\methodabbrev}{\textsc{SkipAt}}

\usepackage[capitalize]{cleveref}
\crefname{section}{Sec.}{Secs.}
\Crefname{section}{Section}{Sections}
\Crefname{table}{Table}{Tables}
\crefname{table}{Tab.}{Tabs.}

\def\cvprPaperID{9811} 
\def\confName{CVPR}
\def\confYear{2023}

\begin{document}

\title{Skip-Attention: Improving Vision Transformers by Paying Less Attention}

\author{Shashanka Venkataramanan$^{1}$\thanks{equal contribution} \thanks{Work done during internship at Qualcomm AI Research} ,\thinspace
Amir Ghodrati$^{1}$\footnotemark[1] ,\thinspace
Yuki M. Asano$^{2}$ \thinspace \\
Fatih Porikli$^{1}$, \thinspace
Amirhossein Habibian$^{1}$ \\
$^{1}${Qualcomm AI Research\thanks{ Qualcomm AI Research is an initiative of Qualcomm Technologies, Inc}} \space,
$^{2}${QUVA Lab, University of Amsterdam}\\
{\tt\small shashanka.venkataramanan@inria.fr \space ghodrati@qti.qualcomm.com
}
}

\maketitle


\newcommand{\head}[1]{{\smallskip\noindent\textbf{#1}}}
\newcommand{\alert}[1]{{\color{red}{#1}}}
\newcommand{\sm}{\scriptsize}
\newcommand{\eq}[1]{(\ref{eq:#1})}

\newcommand{\Th}[1]{\textsc{#1}}
\newcommand{\mr}[2]{\multirow{#1}{*}{#2}}
\newcommand{\mc}[2]{\multicolumn{#1}{c}{#2}}
\newcommand{\tb}[1]{\textbf{#1}}
\newcommand{\ul}[1]{\underline{#1}}
\newcommand{\ch}{\checkmark}

\newcommand{\red}[1]{{\color{red}{#1}}}
\newcommand{\blue}[1]{{\color{blue}{#1}}}
\newcommand{\green}[1]{\color{green}{#1}}
\newcommand{\gray}[1]{{\color{gray}{#1}}}

\newcommand{\citeme}[1]{\red{[XX]}}
\newcommand{\refme}[1]{\red{(XX)}}

\newcommand{\fig}[2][1]{\includegraphics[width=#1\linewidth]{fig/#2}}
\newcommand{\figh}[2][1]{\includegraphics[height=#1\linewidth]{fig/#2}}


\newcommand{\tran}{^\top}
\newcommand{\mtran}{^{-\top}}
\newcommand{\zcol}{\mathbf{0}}
\newcommand{\zrow}{\zcol\tran}

\newcommand{\ind}{\mathbbm{1}}
\newcommand{\expect}{\mathbb{E}}
\newcommand{\nat}{\mathbb{N}}
\newcommand{\zahl}{\mathbb{Z}}
\newcommand{\real}{\mathbb{R}}
\newcommand{\proj}{\mathbb{P}}
\newcommand{\prob}{\mathbf{Pr}}
\newcommand{\normal}{\mathcal{N}}

\newcommand{\mif}{\textrm{if}\ }
\newcommand{\other}{\textrm{otherwise}}
\newcommand{\minimize}{\textrm{minimize}\ }
\newcommand{\maximize}{\textrm{maximize}\ }
\newcommand{\st}{\textrm{subject\ to}\ }

\newcommand{\id}{\operatorname{id}}
\newcommand{\const}{\operatorname{const}}
\newcommand{\sgn}{\operatorname{sgn}}
\newcommand{\var}{\operatorname{Var}}
\newcommand{\mean}{\operatorname{mean}}
\newcommand{\trace}{\operatorname{tr}}
\newcommand{\diag}{\operatorname{diag}}
\newcommand{\vect}{\operatorname{vec}}
\newcommand{\cov}{\operatorname{cov}}
\newcommand{\sign}{\operatorname{sign}}
\newcommand{\prj}{\operatorname{proj}}

\newcommand{\softmax}{\operatorname{softmax}}
\newcommand{\clip}{\operatorname{clip}}

\newcommand{\defn}{\mathrel{:=}}
\newcommand{\peq}{\mathrel{+\!=}}
\newcommand{\meq}{\mathrel{-\!=}}

\newcommand{\floor}[1]{\left\lfloor{#1}\right\rfloor}
\newcommand{\ceil}[1]{\left\lceil{#1}\right\rceil}
\newcommand{\inner}[1]{\left\langle{#1}\right\rangle}
\newcommand{\norm}[1]{\left\|{#1}\right\|}
\newcommand{\abs}[1]{\left|{#1}\right|}
\newcommand{\frob}[1]{\norm{#1}_F}
\newcommand{\card}[1]{\left|{#1}\right|\xspace}
\newcommand{\divg}[2]{{#1\ ||\ #2}}
\newcommand{\diff}{\mathrm{d}}
\newcommand{\der}[3][]{\frac{d^{#1}#2}{d#3^{#1}}}
\newcommand{\pder}[3][]{\frac{\partial^{#1}{#2}}{\partial{#3^{#1}}}}
\newcommand{\ipder}[3][]{\partial^{#1}{#2}/\partial{#3^{#1}}}
\newcommand{\dder}[3]{\frac{\partial^2{#1}}{\partial{#2}\partial{#3}}}

\newcommand{\wb}[1]{\overline{#1}}
\newcommand{\wt}[1]{\widetilde{#1}}

\def\xssp{\hspace{1pt}}
\def\ssp{\hspace{3pt}}
\def\msp{\hspace{5pt}}
\def\lsp{\hspace{12pt}}

\newcommand{\cA}{\mathcal{A}}
\newcommand{\cB}{\mathcal{B}}
\newcommand{\cC}{\mathcal{C}}
\newcommand{\cD}{\mathcal{D}}
\newcommand{\cE}{\mathcal{E}}
\newcommand{\cF}{\mathcal{F}}
\newcommand{\cG}{\mathcal{G}}
\newcommand{\cH}{\mathcal{H}}
\newcommand{\cI}{\mathcal{I}}
\newcommand{\cJ}{\mathcal{J}}
\newcommand{\cK}{\mathcal{K}}
\newcommand{\cL}{\mathcal{L}}
\newcommand{\cM}{\mathcal{M}}
\newcommand{\cN}{\mathcal{N}}
\newcommand{\cO}{\mathcal{O}}
\newcommand{\cP}{\mathcal{P}}
\newcommand{\cQ}{\mathcal{Q}}
\newcommand{\cR}{\mathcal{R}}
\newcommand{\cS}{\mathcal{S}}
\newcommand{\cT}{\mathcal{T}}
\newcommand{\cU}{\mathcal{U}}
\newcommand{\cV}{\mathcal{V}}
\newcommand{\cW}{\mathcal{W}}
\newcommand{\cX}{\mathcal{X}}
\newcommand{\cY}{\mathcal{Y}}
\newcommand{\cZ}{\mathcal{Z}}

\newcommand{\vA}{\mathbf{A}}
\newcommand{\vB}{\mathbf{B}}
\newcommand{\vC}{\mathbf{C}}
\newcommand{\vD}{\mathbf{D}}
\newcommand{\vE}{\mathbf{E}}
\newcommand{\vF}{\mathbf{F}}
\newcommand{\vG}{\mathbf{G}}
\newcommand{\vH}{\mathbf{H}}
\newcommand{\vI}{\mathbf{I}}
\newcommand{\vJ}{\mathbf{J}}
\newcommand{\vK}{\mathbf{K}}
\newcommand{\vL}{\mathbf{L}}
\newcommand{\vM}{\mathbf{M}}
\newcommand{\vN}{\mathbf{N}}
\newcommand{\vO}{\mathbf{O}}
\newcommand{\vP}{\mathbf{P}}
\newcommand{\vQ}{\mathbf{Q}}
\newcommand{\vR}{\mathbf{R}}
\newcommand{\vS}{\mathbf{S}}
\newcommand{\vT}{\mathbf{T}}
\newcommand{\vU}{\mathbf{U}}
\newcommand{\vV}{\mathbf{V}}
\newcommand{\vW}{\mathbf{W}}
\newcommand{\vX}{\mathbf{X}}
\newcommand{\vY}{\mathbf{Y}}
\newcommand{\vZ}{\mathbf{Z}}

\newcommand{\va}{\mathbf{a}}
\newcommand{\vb}{\mathbf{b}}
\newcommand{\vc}{\mathbf{c}}
\newcommand{\vd}{\mathbf{d}}
\newcommand{\ve}{\mathbf{e}}
\newcommand{\vf}{\mathbf{f}}
\newcommand{\vg}{\mathbf{g}}
\newcommand{\vh}{\mathbf{h}}
\newcommand{\vi}{\mathbf{i}}
\newcommand{\vj}{\mathbf{j}}
\newcommand{\vk}{\mathbf{k}}
\newcommand{\vl}{\mathbf{l}}
\newcommand{\vm}{\mathbf{m}}
\newcommand{\vn}{\mathbf{n}}
\newcommand{\vo}{\mathbf{o}}
\newcommand{\vp}{\mathbf{p}}
\newcommand{\vq}{\mathbf{q}}
\newcommand{\vr}{\mathbf{r}}
\newcommand{\Vs}{\mathbf{s}}
\newcommand{\vt}{\mathbf{t}}
\newcommand{\vu}{\mathbf{u}}
\newcommand{\vv}{\mathbf{v}}
\newcommand{\vw}{\mathbf{w}}
\newcommand{\vx}{\mathbf{x}}
\newcommand{\vy}{\mathbf{y}}
\newcommand{\vz}{\mathbf{z}}

\newcommand{\vone}{\mathbf{1}}
\newcommand{\vzero}{\mathbf{0}}

\newcommand{\valpha}{{\boldsymbol{\alpha}}}
\newcommand{\vbeta}{{\boldsymbol{\beta}}}
\newcommand{\vgamma}{{\boldsymbol{\gamma}}}
\newcommand{\vdelta}{{\boldsymbol{\delta}}}
\newcommand{\vepsilon}{{\boldsymbol{\epsilon}}}
\newcommand{\vzeta}{{\boldsymbol{\zeta}}}
\newcommand{\veta}{{\boldsymbol{\eta}}}
\newcommand{\vtheta}{{\boldsymbol{\theta}}}
\newcommand{\viota}{{\boldsymbol{\iota}}}
\newcommand{\vkappa}{{\boldsymbol{\kappa}}}
\newcommand{\vlambda}{{\boldsymbol{\lambda}}}
\newcommand{\vmu}{{\boldsymbol{\mu}}}
\newcommand{\vnu}{{\boldsymbol{\nu}}}
\newcommand{\vxi}{{\boldsymbol{\xi}}}
\newcommand{\vomikron}{{\boldsymbol{\omikron}}}
\newcommand{\vpi}{{\boldsymbol{\pi}}}
\newcommand{\vrho}{{\boldsymbol{\rho}}}
\newcommand{\vsigma}{{\boldsymbol{\sigma}}}
\newcommand{\vtau}{{\boldsymbol{\tau}}}
\newcommand{\vupsilon}{{\boldsymbol{\upsilon}}}
\newcommand{\vphi}{{\boldsymbol{\phi}}}
\newcommand{\vchi}{{\boldsymbol{\chi}}}
\newcommand{\vpsi}{{\boldsymbol{\psi}}}
\newcommand{\vomega}{{\boldsymbol{\omega}}}

\newcommand{\rLambda}{\mathrm{\Lambda}}
\newcommand{\rSigma}{\mathrm{\Sigma}}

\newcommand{\vLambda}{\bm{\rLambda}}
\newcommand{\vSigma}{\bm{\rSigma}}


\makeatletter
\newcommand{\vast}[1]{\bBigg@{#1}}
\makeatother

\makeatletter
\newcommand*\bdot{\mathpalette\bdot@{.7}}
\newcommand*\bdot@[2]{\mathbin{\vcenter{\hbox{\scalebox{#2}{$\m@th#1\bullet$}}}}}
\makeatother

\makeatletter
\DeclareRobustCommand\onedot{\futurelet\@let@token\@onedot}
\def\@onedot{\ifx\@let@token.\else.\null\fi\xspace}

\def\eg{\emph{e.g}\onedot} \def\Eg{\emph{E.g}\onedot}
\def\ie{\emph{i.e}\onedot} \def\Ie{\emph{I.e}\onedot}
\def\cf{\emph{cf}\onedot} \def\Cf{\emph{Cf}\onedot}
\def\etc{\emph{etc}\onedot} \def\vs{\emph{vs}\onedot}
\def\wrt{w.r.t\onedot} \def\dof{d.o.f\onedot} \def\aka{a.k.a\onedot}
\def\etal{\emph{et al}\onedot}
\makeatother

\definecolor{LightCyan}{rgb}{0.88,1,1}

\newcommand{\CC}[1]{\cellcolor{LightCyan!#1}}

\newcommand{\cmark}{\ding{51}}%
\newcommand{\xmark}{\ding{55}}%

\begin{strip}\centering
\includegraphics[width=\textwidth]{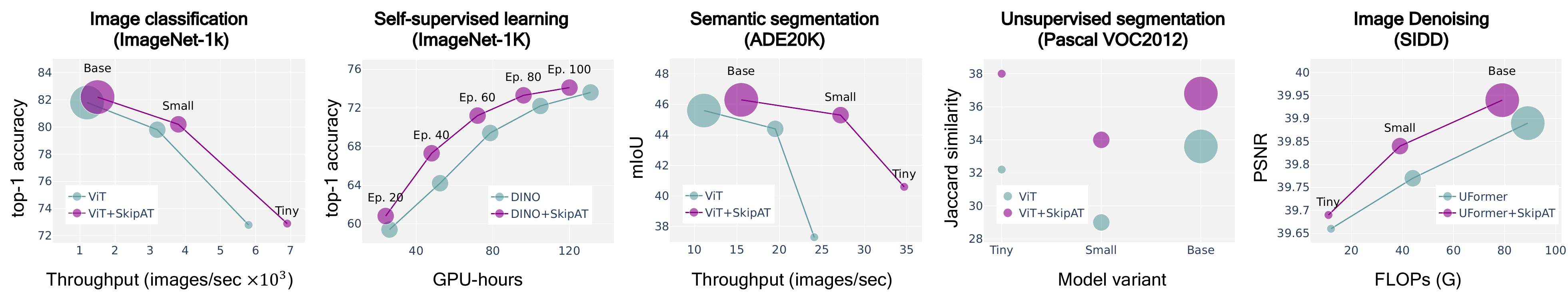}
\captionof{figure}{\tb{Performance of~\methodabbrev~across 5 different tasks.} Our novel~\methodabbrev~parametric function achieves superior accuracy \vs efficiency trade-off over the baseline transformer on a wide array of tasks.}
\label{fig:teaser}
\end{strip}

\begin{abstract}
This work aims to improve the efficiency of vision transformers (ViT). 
While ViTs use computationally expensive self-attention operations in every layer, we identify that these operations are highly correlated across layers --  a key redundancy that causes unnecessary computations. Based on this observation, we propose~\methodabbrev, a method to reuse self-attention computation from preceding layers to approximate attention at one or more subsequent layers. To ensure that reusing self-attention blocks across layers does not degrade the performance, we introduce a simple parametric function, which outperforms the baseline transformer's performance while running computationally faster. We show the effectiveness of our method in image classification and self-supervised learning on ImageNet-1K,  semantic segmentation on ADE20K, image denoising on SIDD, and video denoising on DAVIS. We achieve improved throughput at the same-or-higher accuracy levels in all these tasks.

\end{abstract}
\setlength{\parindent}{0pt}
\section{Introduction}
\label{sec:introduction}


The transformer architecture~\cite{vaswani2017attention} has become an important and highly influential model family, due to its simplicity, scalability, and its wide range of applications.
While originally stemming from the domain of natural language processing (NLP), with the advent of the Vision transformer (ViT)~\cite{dosovitskiy2020image}, this has become a standard architecture in computer vision, setting various state-of-the-art (SoTA) performances on tasks ranging from representation learning, semantic segmentation, object detection and video understanding~\cite{caron2021emerging, liu2021swin, carion2020end, liang2022vrt, girdhar2019video}.

However, the original formulation of the transformer includes a quadratic computational complexity with respect to the number of input tokens. 
Given that this number typically ranges from $14^2$ for image classification all the way to $128^2=~16$K for image denoising, this constraint on memory and compute severely limits its applicability.
To tackle this problem, there have been three sets of approaches.
The first leverages redundancies across input tokens and simply reduces computation by efficient sampling, \eg, dropping or merging redundant tokens~\cite{tang2022patch, fayyazadaptive, yin2022vit}.
This, however, means that the final output of the ViT is not spatially continuous and can thus not be used beyond image-level applications such as semantic segmentation or object localization. 
The second set of approaches aims to cheaply estimate the attention computation, but generally at the cost of reduced performances~\cite{yu2022metaformer, chu2021twins}.
Finally, another line of works aims to merge convolutional architectures with the transformer, yielding hybrid architectures~\cite{li2021uniformer, li2021uniformer, pan2022edgevits}. While these increase speed, they do not tackle the fundamental problem of the quadratic complexity, and often introduce an exorbitant number of design choices (essentially a union of those of the transformer and CNNs).

In this work, we propose a novel, so far unexplored approach to solving this problem: simply approximating the computationally expensive blocks of the transformer with a much faster, simpler parametric function.
To arrive at this solution, we first thoroughly analyse the crucial multi-head self-attention (MSA) block of the ViT. 
Through this analysis, we 
find that the attention of the \texttt{CLS} tokens to the spatial patches has a very high correlation across the transformer's blocks, thus leading to unnecessary computations. 
This motivates our approach to leverage attention from an early part of the model and simply reuse it for deeper blocks -- basically ``skipping'' subsequent SA calculations instead of recomputing them  at every layer.

Based on this, we go one step further and explore if the \textit{entire} MSA block of a layer can be skipped by reusing the representation from previous layers.
We find that a simple parametric function inspired from ResneXt's depth-wise convolutions~\cite{xie2017aggregated} can outperform the baseline performance -- while being computationally faster in terms of throughput and FLOPs.
Our method is general-purpose and can be applied to a ViT in any context: \autoref{fig:teaser} shows that our novel parametric function for Skipping Attention (\methodabbrev) achieves superior accuracy \vs efficiency trade-off compared to the baseline transformer on a wide variety of tasks, datasets and model sizes.


In summary, our main contributions are as follows:
\begin{enumerate}[itemsep=4pt, parsep=0pt, topsep=0pt,labelwidth=20pt,leftmargin=20pt]
    \item We propose a novel plug-in module that can be placed in any ViT architecture for reducing the costly $\mathcal{O}(n^2)$ Self-Attention computations (\autoref{sec:param-prop})
    \item We achieve state-of-the-art performances in terms of throughput at same-or-better accuracies for ImageNet, Pascal-VOC2012, SIDD, DAVIS and ADE20K (in the latter of which we obtain 40\% speedup) (\autoref{sec:experiments})
    \item We further demonstrate the generality of our method by obtaining a 26\% reduction in self-supervised pretraining time (at no downstream accuracy loss)  and by demonstrating superior on-device latency (\autoref{sec:exp-dino}, \autoref{sec:exp-imagenet})  
    \item Finally, we analyse the sources of performance gains and extensively ablate our method to provide a model family which can be used for trading off accuracy and throughput (\autoref{sec:ablations}) 
\end{enumerate}

\begin{figure*}[h]
\centering
\includegraphics[width=1\linewidth]{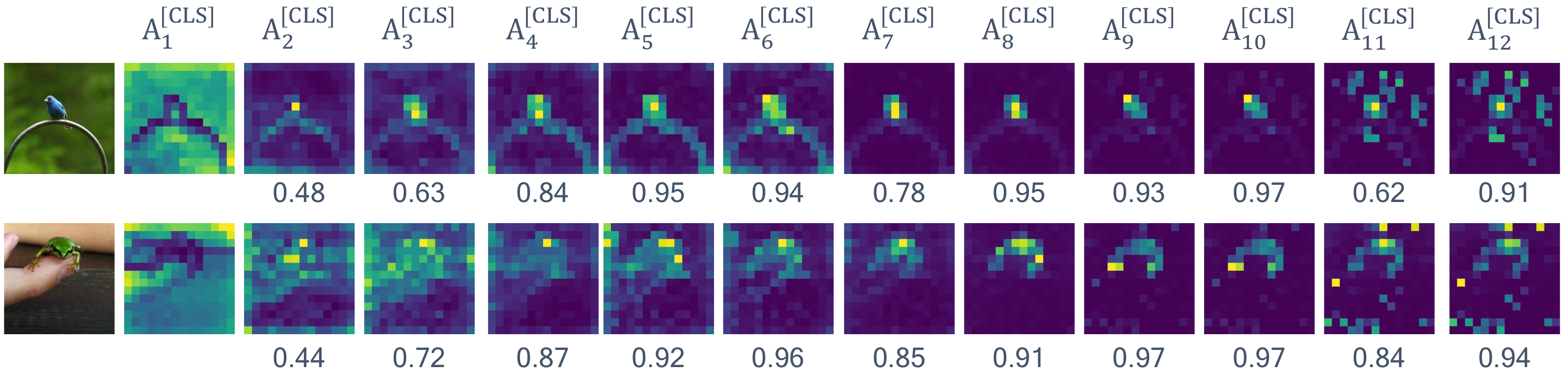}
\caption{\textbf{Attention correlation}. Mean of the attention heads from the \texttt{CLS} token of a pretrained ViT-T/16 at different layers from the validation set of ImageNet-1K. Numbers below each attention map indicates the cosine similarity of $A_{l}^{\texttt{[CLS]}}$ with $A_{l-1}^{\texttt{[CLS]}}$.} \label{fig:attn_corr}
\end{figure*}

\begin{figure}[h]
\centering
\includegraphics[width=1\linewidth]{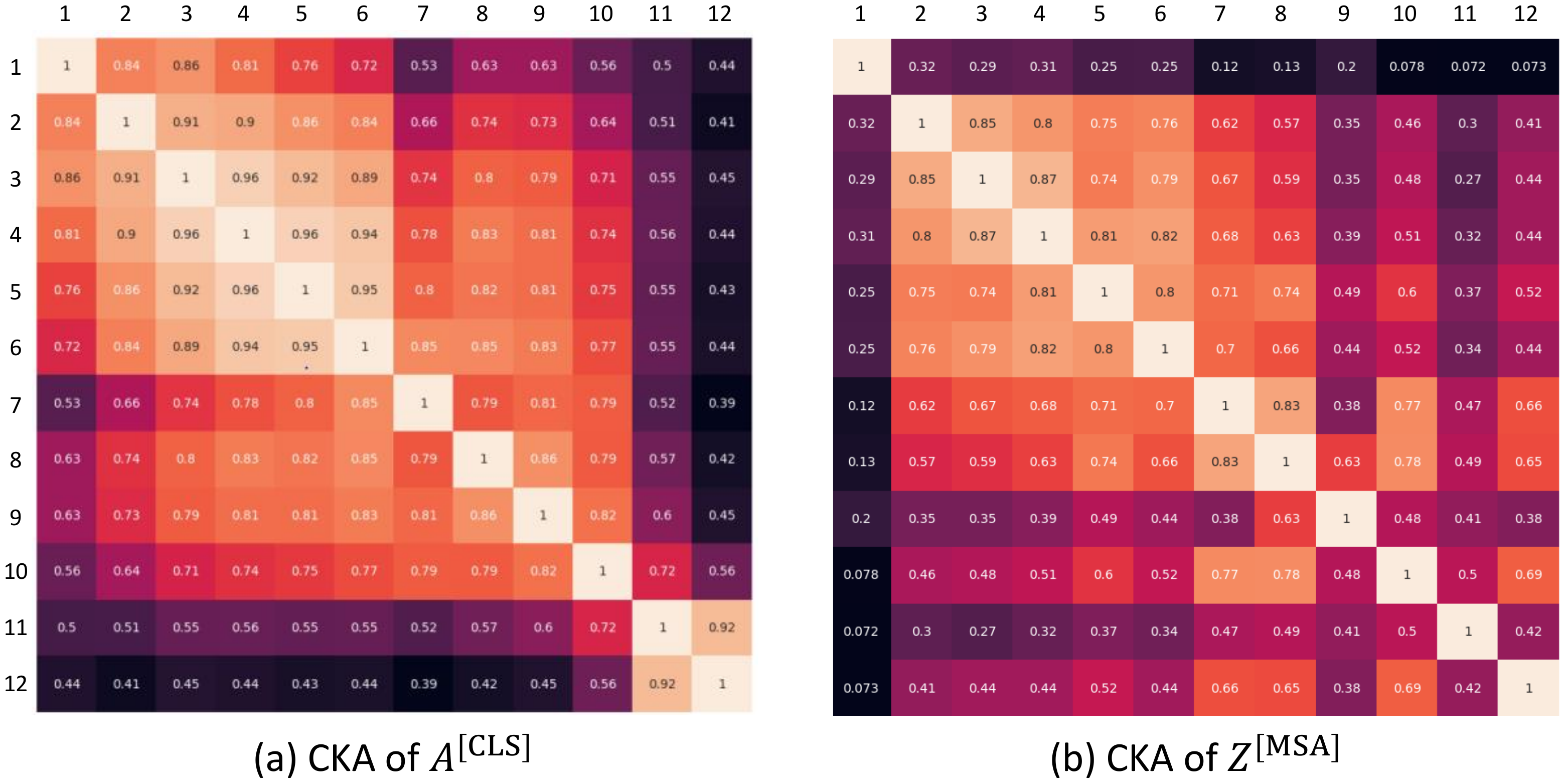}
\caption{\tb{CKA analysis of $A^{[\texttt{CLS}]}$ and $Z^{\text{MSA}}$} across different layers of pretrained ViT-T/16 on the validation set of Imagenet-1K. Vanilla ViT-T/16 has high correlation across both attention maps (layer 3 to 10) and $Z^{\text{MSA}}$ (layer 2 to 8)}
\label{fig:feat_CKA}
\end{figure}
\section{Related Work}
\label{sec:related_work}
There has been great effort made to improve the efficiency of vision transformers (ViT)~\cite{dosovitskiy2020image} from multiple aspects:

\paragraph{Token sampling} 
improves the efficiency either by restructuring images during the tokenization step~\cite{yuan2021tokens, han2021transformer}, pruning the redundant tokens over training~\cite{kong2021spvit, tang2022patch} or dynamically at inference~\cite{yin2022vit, rao2021dynamicvit, fayyazadaptive, chen2021chasing}. Despite their effectiveness in reducing the computational cost in image classification, token sampling methods are hardly applicable to dense prediction tasks,~\eg semantic segmentation and image denoising, where the output image should be spatially continuous. Our approach is complementary to these lines of work and performs favorably against them as validated experimentally. Moreover, given that we keep representing all tokens throughout the network, our approach is applicable to both classification and dense prediction tasks.

\paragraph{Hybrid architectures} 
integrate efficient convolutional modules into vision transformers~\cite{liu2022convnet, pan2022edgevits, mehta2022separable} by adoption of MobileNet blocks in Uniformer~\cite{li2021uniformer}, MobileNetV2 blocks in MobileViT~\cite{mehta2021mobilevit} or using stacks of convolutions in the image tokenization step~\cite{graham2021levit, wu2021cvt}. Similarly, we use convolutions to speed up vision transformers, however, instead of crafting customized blocks as in~\cite{mehta2021mobilevit, pan2022edgevits, mehta2022separable, li2021uniformer}, we adhere to the original transformer architecture and approximate entire MSA computations through convolutions.

\paragraph{Efficient attentions} 
address the quadratic cost of the self-attention operation in vision transformers by global down-sampling of key and value embeddings~\cite{wang2021pyramid, wu2021cvt}, performing self-attention in local windows~\cite{liu2021swin}, alternating between local and global self-attentions~\cite{chu2021twins, mehta2021mobilevit, pan2022edgevits}, or replacing self-attention with a simple pooling~\cite{yu2022metaformer}. However, reducing the self-attention to a local neighborhood hinders their ability to model the long range dependencies and leads to a significant performance drop with moderate speed up~\cite{zhang2021multi}. Moreover, some of the introduced operations come with no efficient support,~\eg cyclic shift in Swin~\cite{liu2021swin}, limiting their actual efficiency gains in terms of latency. Different to this, our method relies on the strong, yet inefficient self-attention operator at a few blocks and lighter, accurate attention estimators in other blocks. As the estimators only rely on standard convolutional operations, our method translates to actual latency gains. Related to this paper,~\cite{xiao2019sharing, wang2021evolving, ying2021lazyformer} observed the redundancies in attention maps, for NLP tasks. However, instead of simply copying attention maps~\cite{xiao2019sharing, ying2021lazyformer}, we propose an efficient parametric function that, as we show, are critical to achieve a high throughput whilst retaining high model performance in vision tasks.

\paragraph{Hierarchical architectures} 
introduce hierarchical representations, as a long-standing principle in computer vision, to vision transformers~\cite{wang2021pyramid, graham2021levit, liu2021swin, zhang2021multi, pan2021scalable}. Using a multi-scale representation significantly improves the memory and computational cost of the isotropic architectures, such as ViT. More recently, the idea has been extended to more complex architectures with U-Net~\cite{wang2022uformer} or multi-branch structures~\cite{gu2022multi}. Our work is complementary to these works, as they do not tackle the fundamental problem of reducing the quadratic complexity of the self-attention operator. We experimentally validate the effectiveness of our method on such isotropic and hierarchical architectures.

\section{\methodname}
\subsection{Preliminaries}
\label{sec: prelim}
\paragraph{Vision Transformer.}
Let $x \in \real^{h \times w \times c}$ be an input image, where $h \times w$ is the spatial resolution and $c$ is the number of channels.
The image is first tokenized into $n = hw/p^2$ non-overlapping patches, where $p \times p$ is patch size. Each patch is projected into an embedding $z_i \in \real^d$ using a linear layer 
to obtain the tokenized image:
\begin{align}
    Z_0 = (z_1; \dots; z_n) \in \real^{n \times d}
    \label{eq:token-embed}
\end{align}
Here, $``;"$ denotes row-wise stacking. Positional embeddings are added to $Z_0$ to retain positional information. The token embeddings are then input to a $\cL = \{1,\dots,L\}$ layer transformer whose output is denoted as $Z_L$. 
In the supervised setting, a learnable token $z^{[\texttt{CLS}]} \in \real^d$ is prepended to the tokenized image in ~\eqref{eq:token-embed} as $Z_0 \defn (z^{[\texttt{CLS}]};Z_0) \in \real^{(n+1) \times d}$.

\textbf{Transformer Layer.}
Every layer of the transformer consists of a multi-head self attention (MSA) block followed by a multi-layer perceptron (MLP) block. 
In the MSA block, the input, $Z_{l-1} \in \real^{n \times d}$, for $l \in \cL$, is first projected into three learnable embeddings  $\{Q,K,V\} \in \real^{n \times d}$. The attention matrix $A$, is calculated as 
\begin{align}
    A \defn \sigma \left(\frac{Q K^T}{\sqrt{d}} \right) \in \real^{n \times n}
    \label{eq:attn}
\end{align}
where $\sigma(.)$ denotes the row-wise softmax operation. The ``multi-head" in MSA is defined by considering $h$ attention heads where each head is a sequence of $n \times \frac{d}{h}$ matrix.
The attention heads are reprojected back to $n \times d$ using a linear layer which is combined with the value matrix as
\begin{align}
    Z^{\text{MSA}} \defn AV \in \real^{n \times d}
    \label{eq:val-feats}
\end{align}
The output representations from the MSA block is then input to the MLP block which comprises two linear layers separated by a GeLU activation~\cite{hendrycks2016gaussian}. 
At a given layer $l$, the computational flow of representations through a transformer block is denoted as 
\begin{align}
    Z_{l} &\gets  Z_{l}^{\text{MSA}} + Z_{l-1} \label{eq:msa-flow},\\
    Z_{l} &\gets \text{MLP}(Z_l) + Z_{l} \label{eq:mlp-flow}.
\end{align}
Both the MSA and MLP blocks have residual connections with layer normalization (LN)~\cite{ba2016layer}. 
While MSA blocks in each layer of the transformer learn representations independently, in the next subsection, we show that empirically there exist high correlation across these layers.

\begin{figure*}
\centering
\includegraphics[width=1\linewidth]{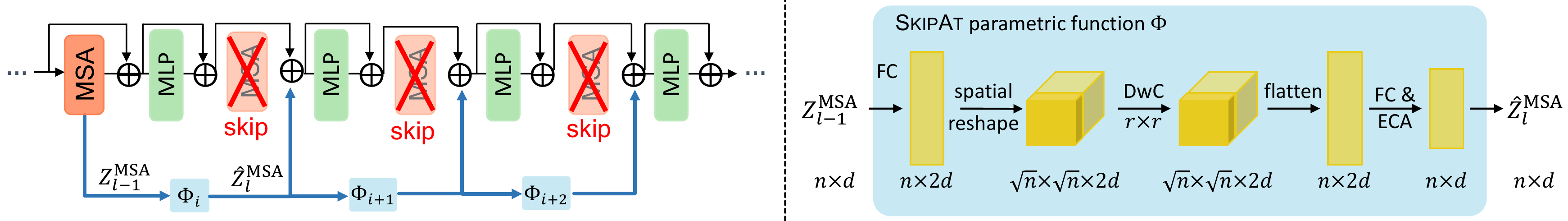}
\caption{\tb{\methodabbrev~framework} We illustrate \methodabbrev~on ViT~\cite{dosovitskiy2020image}. The \methodabbrev~ parametric function ($\Phi$) uses representations of the MSA block (in solid color) $Z_{l-1}^{\text{MSA}}$ as input, which undergoes a series of transformations. An element-wise summation ($\bigoplus$) with the output of the MLP block from layer $l-1$ and $\hat{Z}_{l}^{\text{MSA}}$ is used as input to the MLP block at layer $l$. The MSA operation (crossed out) is thus not computed and is discarded from the computational graph. With \methodabbrev~the total number of layers remains unchanged.
}
\label{fig:framework}
\vspace{-6pt}
\end{figure*}

\subsection{Motivation: Layer Correlation Analysis}
\label{sec:feat-corr}

\paragraph{Attention-map correlation.}
The MSA block in ViT encodes the similarity of each patch to every other patch as an $n \times n$ attention matrix. This operator is computationally expensive with $\cO(n^2)$ complexity \eqref{eq:attn}. As ViTs scale up, \ie, as $n$ increases, the complexity grows quadratically and this operation becomes a bottleneck. Recent NLP works~\cite{vig2019analyzing, vig2019multiscale} have shown that self-attention across adjacent layers in SoTA language models exhibit very high correlation. This raises the question --~\emph{is it worth to compute self-attention at every layer of a vision transformer?} 

To address this question, we analyze the correlation of the self-attention maps across different layers of ViT. 
As shown in~\autoref{fig:attn_corr}, the self-attention maps from the class token, $A^{\texttt{[CLS]}}$, exhibit high correlation especially in the intermediate layers. 
The cosine similarity between $A_{l-1}^{\texttt{[CLS]}}$ and $A_{l}^{\texttt{[CLS]}}$ can be as high as $0.97$, as indicated in the bottom of each attention map in~\autoref{fig:attn_corr}. Similar behavior is observed from other token embeddings, which we analyze in the supplementary material. We quantitatively analyze this correlation across all the samples of the validation set of ImageNet-1K, by computing the Centered Kernel Alignment (CKA)~\cite{kornblith2019similarity, cortes2012algorithms} between $A_{i}^{\texttt{[CLS]}}$ and $A_{j}^{\texttt{[CLS]}}$ for every $i,j \in \cL$. CKA measures the similarity between representations obtained from intermediate layers of the network, where a high value of CKA indicates high correlation between the representations. From~\autoref{fig:feat_CKA} (a), we observe that ViT-T has a high correlation across $A^{\texttt{[CLS]}}$ especially from layer 3 through 10.

\paragraph{Feature correlation.} In ViTs, the high correlation is not just limited to $A^{\texttt{[CLS]}}$, but the representation from MSA blocks, $Z^{\text{MSA}}$, also show high correlation throughout the model~\cite{raghu2021vision}. To analyze the similarity across these representations, we compute the CKA between $Z^{\text{MSA}}_i$ and $Z^{\text{MSA}}_j$ for every $i,j \in \mathcal{L}$. We observe from \autoref{fig:feat_CKA} (b), that $Z^{\text{MSA}}$ also have high similarity across adjacent layers of the model especially in the earlier layers, \ie, from layer 2 through 8.  

\subsection{Improving Efficiency by Skipping Attention}
\label{sec:param-prop}
Based on our observation of high representation similarity across MSA blocks of a transformer (\autoref{sec:feat-corr}), we propose to leverage the correlation across both the attention matrix and the representations from the MSA block to improve the efficiency of vision transformers. Instead of computing the MSA operation \eqref{eq:val-feats} independently at every layer, we explore a simple and effective strategy to utilize dependencies across the features from these layers. 

In particular, we propose to skip MSA computation in one or more layers of a transformer by reusing representations from its adjacent layers. 
We term this operation as \emph{Skip Attention} or \emph{~\methodabbrev}. As the compute and memory benefit from skipping the entire MSA block is greater than skipping just the self-attention operation ($\cO(n^2d + nd^2)$ \vs $\cO(n^2d)$), in this paper we focus on former.  
However, instead of directly re-using features, \ie, copying the features from the source MSA block to one or more adjacent MSA blocks, we introduce a parametric function. The parametric function ensures that directly reusing features does not affect the translation invariance and equivariance in these MSA blocks and acts as a strong regularizer to improve model generalization. 

\paragraph{\methodabbrev~parametric function} Let $\Phi: \real^{n \times d} \to \real^{n \times d}$ denote the parametric function that maps output of the MSA block from $l-1$ to $l$ as $\hat{Z}^{\text{MSA}}_{l} \defn \Phi(Z^{\text{MSA}}_{l-1})$. Here, $\hat{Z}^{\text{MSA}}_{l}$ is the approximation of $Z^{\text{MSA}}_{l}$. The parametric function can be as simple as an identity function, where $Z_{l-1}^{\text{MSA}}$ is directly reused. Instead of computing MSA operation at $l$, we use $Z^{\text{MSA}}_{l-1}$ as the input to the MLP block at $l$. 
%
When using an identity function, due to the absence of MSA operation at $l$, the relation across tokens is no longer encoded in the attention matrix, which affects representation learning. To mitigate this, we introduce the \methodabbrev~parametric function inspired from ResNeXt~\cite{xie2017aggregated} as shown in~\autoref{fig:framework}, to encode local relations among tokens. The \methodabbrev~parametric function consists of two linear layers and a depth-wise convolution (DwC)~\cite{chollet2017xception} in between, as follows:

\begin{align}
    \hat{Z}_{l}^{\text{MSA}}  \defn \text{ECA} \Bigl (\text{FC}_2 \Bigl ( \text{DwC} \bigl (\text{FC}_1 (Z_{l-1}^{\text{MSA}}) \bigr ) \Bigr )\Bigr )
\end{align}

In the case of supervised learning, we first separate the \texttt{CLS} embeddings from $Z^{\text{MSA}} \in \real^{(n+1) \times d}$ into class embeddings $Z_C^{\text{MSA}} \in \real^d$ and the patch embeddings to $Z_{P}^{\text{MSA}} \in \real^{n \times d}$. The patch embeddings are then input to the first linear layer $\text{FC}_1: \real^{n \times d} \to \real^{n \times 2d}$, which expands the channel dimension. This is followed by $\text{DwC}: \real^{\sqrt{n} \times \sqrt{n} \times 2d} \to \real^{\sqrt{n} \times \sqrt{n} \times 2d}$ with kernel $r \times r$ to capture cross-token relations. Note that before the DwC operation, we spatially reshape the input matrix to a feature tensor. The output of the DwC is then flattened back to a vector and fed to the last FC layer $\text{FC}_2:  \real^{n \times 2d} \to \real^{n \times d}$ which reduces the channel dimension back to its initial dimension $d$. We use GeLU activations after $\text{FC}_1$ and $\text{DwC}$. Following~\cite{wang2020eca}, we use efficient channel attention module (ECA) after $\text{FC}_2$ to enhance the cross-channel dependencies. The ECA module first aggregates the features along the channel dimension using global average pooling (GAP). A $1 \times 1$ convolution with adaptive kernel size proportional to channel dimension is applied followed by sigmoid activation. This operation of the ECA module enhances cross-channel dependencies. We then concatenate the embedding of the class-token with the output of the ECA to obtain $\hat{Z}_{l}^{\text{MSA}}$.

\paragraph{\methodabbrev~framework.}
The overall framework of \methodabbrev~is illustrated in \autoref{fig:framework}. ~\methodabbrev~can be incorporated  into any transformer architecture which we empirically show in~\autoref{sec:exp-denoising}. Depending on the architecture, one can skip the MSA operation in one or more layers of the transformer. In ViT, as we empirically observe that representations from the MSA block, $Z^{\text{MSA}}$, have high correlations from layer 2 through 7 (\autoref{sec:feat-corr}), we employ the \methodabbrev~ parametric function in these layers. 
This means that we use the $Z_{2}^{\text{MSA}}$ as input to the \methodabbrev~parametric function and skip MSA operations in layers 3-8. Instead, the features from the output of the \methodabbrev~parametric function is used as input to the MLP block. The computation flow of representations is now modified to
\begin{align}
    Z_{l} &\gets \Phi(Z^{\text{MSA}}_{l-1})+Z_{l-1} \\
    Z_{l} &\gets \text{MLP}(Z_l) + Z_{l}
\end{align}
Due to the presence of residual connections in the MSA and MLP blocks, which is standard in ViT~\cite{dosovitskiy2020image}, the MLP blocks at layer 3 through 8 learn representations independently and cannot be discarded from the computational graph. It is important to note that, with~\methodabbrev~the total number of layers in ViT remain unchanged, but there are fewer MSA blocks.

\paragraph{Complexity: MSA \vs \methodabbrev}
The self-attention operation involves three operations. Firstly, the token embeddings are projected into query, key and value embeddings, secondly, attention matrix $A$ is computed as dot product between $Q$ and $K$ and finally, the output representations are computed as dot product between $A$ and $V$. This results in a complexity of $\cO(4nd^2 + n^2d)$. Since $d \ll n$, the complexity of MSA block can be reduced to $\cO(n^2d)$.

The \methodabbrev~parametric function consists of two linear layers and one depth-wise convolution operation, which results in a $\cO(2nd^2 + r^2nd)$ complexity, where $r \times r$ is the kernel size of the DwC operation. The overall complexity of \methodabbrev~can be reduced to $\cO(nd^2)$ since $r^2 \ll d$. 
Thus, \methodabbrev~has fewer FLOPs than the MSA block as $\cO(nd^2) < \cO(n^2d)$ when $n$ increases as transformers scale up.

\section{Experiments}
\label{sec:experiments}

\subsection{Image Classification}
\label{sec:exp-imagenet}
\begin{table}
\centering
\scriptsize
\setlength{\tabcolsep}{2.5pt}
\begin{tabular}{llcccc} \toprule
 \Th{Backbone}    & \Th{Method}                             &\Th{top-1}$\uparrow$   & \Th{Param}$\downarrow$        & \Th{GFlops}$\downarrow$   & \Th{Throughput}$\uparrow$          \\ 
                  &                                         & (\%)        & ($\times 10^6$)   &               & (\Th{im/s} $\times 10^3$)  \\ \midrule
                  & ViT~\cite{dosovitskiy2020image}         & 72.8        & 5.7               & 1.2           & 5.8                      \\
                  & A-ViT~\cite{yin2022vit}                 & 71.0        & 5.7               & 0.8           & 6.3                      \\
                  & Dynamic ViT~\cite{rao2021dynamicvit}    & 70.9        & --                & 0.9           & 6.1                      \\
                  & SViTE~\cite{chen2021chasing}            & 71.7        & \tb{4.0}          & 0.9           & 6.2                      \\
ViT-T/16          & SPViT~\cite{kong2021spvit}              & 72.7        & 5.7               & 0.9           & 6.7                      \\
                  & ATS~\cite{fayyazadaptive}               & 72.7        & 5.7               & 0.9           & 6.1                      \\
                  & PS-ViT~\cite{tang2022patch}             & 72.6        & --                & \tb{0.7}      & 6.6                      \\
                  & HVT~\cite{pan2021scalable}              & 70.2        & 5.7               & \tb{0.7}      & \tb{7.2}                 \\
                  \cmidrule{2-6}
                  \rowcolor{LightCyan}
                  & ~\methodabbrev                          & \tb{72.9}   & 5.8               & 1.1           & 6.9                      \\
                  \midrule 
                  & ViT~\cite{dosovitskiy2020image}         & 79.8        & 22.4              & 4.6           & 3.2                      \\
                  & A-ViT~\cite{yin2022vit}                 & 78.6        & 22.4              & 3.6           & 3.4                      \\
                  & Dynamic ViT~\cite{rao2021dynamicvit}    & 78.3        & 23.1              & 3.4           & 3.6                      \\
                  & SViTE~\cite{chen2021chasing}            & 80.2        & \tb{13.1}         & 2.7           & 3.5                      \\
ViT-S/16          & ATS~\cite{fayyazadaptive}               & 79.7        & 22.4              & 2.9           & 3.3                      \\
                  & PS-ViT~\cite{tang2022patch}             & 79.4        & --                & 2.6           & 3.9                      \\
                  & SPViT~\cite{kong2021spvit}              & 79.3        & 22.1              & 2.7           & 3.5                      \\
                  & Rev-ViT~\cite{mangalam2022reversible}   & 79.8        & 22.4              & 4.6           & 3.6                      \\
                  & HVT\cite{pan2021scalable}               & 78.0        & 22.5              & \tb{2.4}      & \tb{4.1}                  \\
                  \cmidrule{2-6}
                  \rowcolor{LightCyan}
                  &  ~\methodabbrev                         & \tb{80.2}   & 22.1              & 4.0           & 3.8                      \\
                  \midrule
                  & ViT~\cite{dosovitskiy2020image}         & 81.8        & 87.3              & 17.6          & 1.2                      \\
                  & SViTE~\cite{chen2021chasing}            & 81.6        & \tb{52.0}         & 11.5          & 1.3                         \\
ViT-B/16                  & Rev-ViT~\cite{mangalam2022reversible}   & 81.5        & 87.3              & 17.6          & 1.2                         \\
                  & PS-ViT~\cite{tang2022patch}             & 81.5        &  --               & \tb{9.8}      & \tb{1.6}                         \\
                  \cmidrule{2-6}
                  \rowcolor{LightCyan}
                  &~\methodabbrev                           & \tb{82.2}   & 86.7              & 15.2          & 1.5                       \\
                  \bottomrule
\end{tabular}
\vspace{-6pt}
\caption{\tb{Image classification on ImageNet-1K.} Accuracy \vs efficiency comparison of \methodabbrev~with SoTA methods for image resolution $224 \times 224$. For all the methods, we measure throughput (image/sec) with a batch size of 1024 on a single NVIDIA A100 GPU, averaged over the validation set of ImageNet-1K.}
\label{table:sota-imagenet}
\vspace{-10pt}
\end{table}

We use ViT-T/16~\cite{dosovitskiy2020image}, ViT-S/16~\cite{dosovitskiy2020image} and ViT-B/16~\cite{dosovitskiy2020image} as our backbone on ImageNet-1K. For fair comparisons, we follow the experimental settings in~\cite{touvron2021training} and evaluate \methodabbrev~against SoTA methods: A-ViT~\cite{yin2022vit}, Dynamic-ViT~\cite{mou2021dynamic}, SViTE~\cite{chen2021chasing}, SPViT~\cite{kong2021spvit}, ATS~\cite{fayyazadaptive}, PS-ViT~\cite{tang2022patch}, HVT~\cite{pan2021scalable} and Rev-Vit~\cite{mangalam2022reversible}. To the best of our knowledge, these are all the works that improve the efficiency of ViT without modifying its underlying architecture. 
 
From~\autoref{table:sota-imagenet}, we observe that \methodabbrev~achieves the best accuracy \vs efficiency trade-off compared to all SoTA methods on different variants of ViT. Notably, we outperform baseline ViT-T, ViT-S and ViT-B by 0.1\%, 0.4\% and 0.4\% respectively, while SoTA methods achieve lower accuracy or are on-par with the baseline. Since \methodabbrev~uses a parametric function to skip computing MSA blocks, our reduction in number of parameters and in FLOPs is comparable to the SoTA. In terms of throughput, \methodabbrev~is 19\%, 21\% and 25\% faster than the baseline ViT-T, ViT-S and ViT-B respectively. 
Dehghani \textit{et al.}~\cite{dehghani2021efficiency} highlight the significance of using \emph{throughput} as a metric to measure model efficiency: as the reduction in FLOPs does not necessarily correspond to improvements in latency, as it does not take into account the degree of parallelism or other hardware details. In line with this argument, we observe that while SoTA methods such as ATS~\cite{fayyazadaptive} and SPViT~\cite{kong2021spvit} achieve large reduction in FLOPs, they actually have lower throughput when compared to \methodabbrev. Furthermore, HVT~\cite{pan2021scalable} while achieving a higher gain in both throughput and FLOPs has poor top-1 accuracy (2.6\% drop in ViT-T and 1.8\% drop in ViT-S). Thus, \methodabbrev~demonstrates the ability to simultaneously improve both accuracy and throughput over SoTA methods.
\begin{figure}[t]
\centering
\includegraphics[width=.9\linewidth]{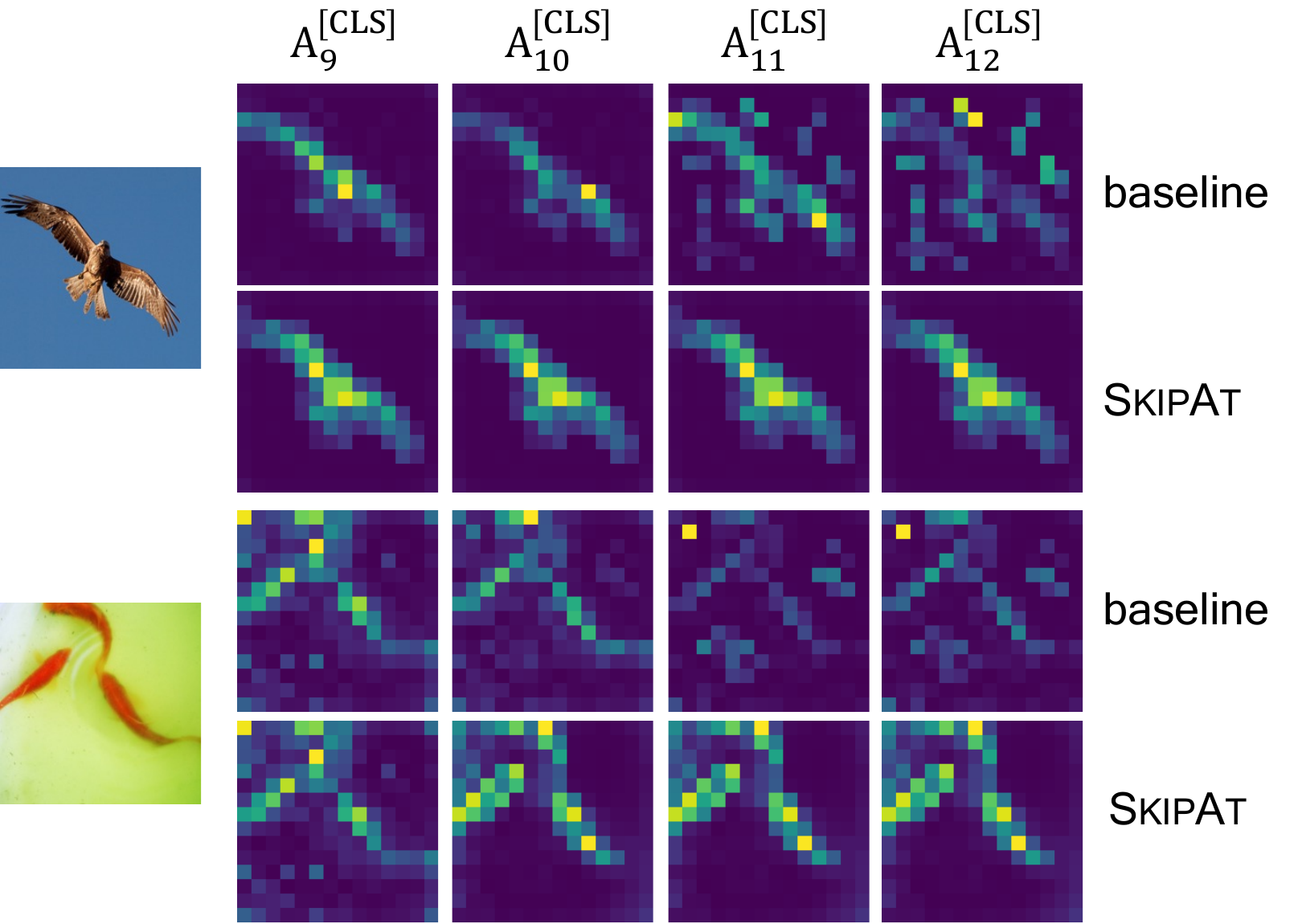}
\caption{\tb{Visualizing attention maps}. Mean of the attention of different heads from $A^{[\texttt{CLS}]}$ from last four layers of ViT-T/16 on the validation set of ImageNet-1K. Attention maps from last four blocks show that~\methodabbrev~localizes the object better than vanilla ViT.}
\label{fig:attn_ours}
\end{figure}

\begin{figure}[t]
\centering
\includegraphics[width=.6\linewidth]{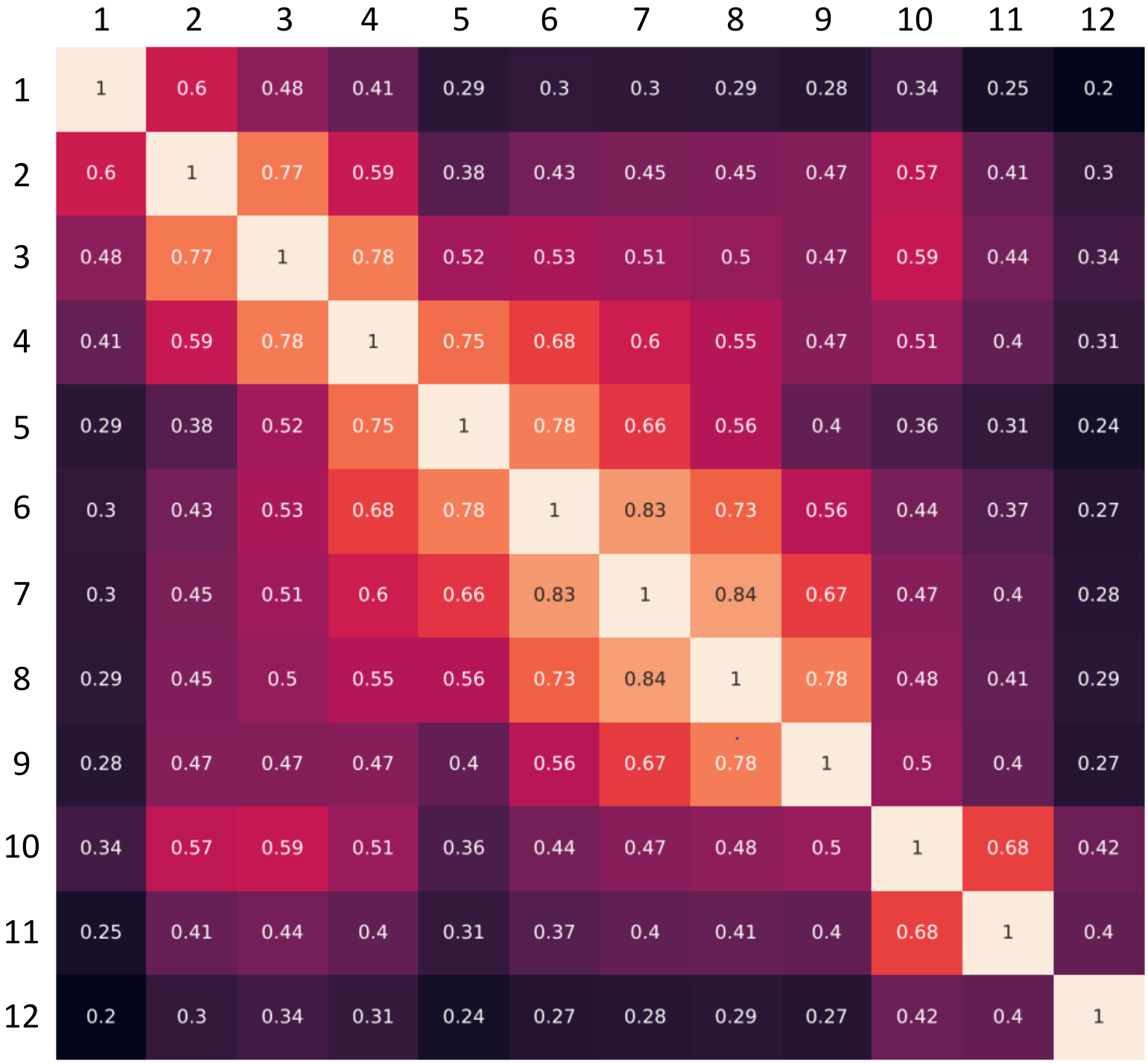}
\caption{\tb{CKA analysis of \methodabbrev} shows that $Z^{\text{MSA}}$ has lower correlation between layers. The high correlation is only between consecutive layers 2 through 8, where the MSA operation is skipped.  }
\label{fig:CKA_ours}
\end{figure}

\textbf{Visualizing attention maps and $Z^{\text{MSA}}$ correlation.}
We analyze the effect of the \methodabbrev~parametric function by visualizing the mean of attention heads of the \texttt{CLS} token from the last four layers of ViT-T/16. From~\autoref{fig:attn_ours}, we observe that while attention maps from vanilla ViT (last two layers) do not solely attend to the object, the attention maps from \methodabbrev~accurately focuses on the object. It is interesting to note that, the attention maps from \methodabbrev~are also capable of attending to multiple objects in the image (\autoref{fig:attn_ours}: second example). We further analyze the CKA of the representations from MSA block across all the layers of ViT-T/16. From ~\autoref{fig:CKA_ours}, we observe that $Z^{\text{MSA}}$ has lower correlation across layers except between the layers where the MSA operation is skipped (layer 3 to 8). However, unlike vanilla ViT (\autoref{fig:feat_CKA} (b)) the correlation from each layer to every other layer is quite low.  This shows that our \methodabbrev~parametric function acts as a strong regularizer and thus improves the representations of the model.
\begin{table}[h]
\centering
\small
\begin{tabular}{cccc}
\toprule
\Th{Method}                        & \Th{Jaccard}$\uparrow$  & \Th{CorLoc}$\uparrow$\\ 
\toprule
ViT-T~\cite{dosovitskiy2020image}  & 32.2          & 39.5      \\
\rowcolor{LightCyan}
ViT-T + \methodabbrev              & \tb{38.0}     & \tb{41.5} \\
\cmidrule{2-3}
ViT-S~\cite{dosovitskiy2020image}  & 29.0          & 40.6      \\
\rowcolor{LightCyan}
ViT-S + \methodabbrev              & \tb{34.0}     & \tb{41.2} \\
\cmidrule{2-3}
ViT-B~\cite{dosovitskiy2020image}  & 33.6          & 36.4      \\
\rowcolor{LightCyan}
ViT-B + \methodabbrev              & \tb{36.8}     & \tb{37.2} \\
 \bottomrule
\end{tabular}
\vspace{-6pt}
\caption{\textbf{Unsupervised Segmentation and Object Localization} using Jaccard similarity~\cite{caron2021emerging} and Correct Localization (CorLoc)~\cite{melas2022deep}, on the validation set of Pascal VOC2012. All models have been pretrained on ImageNet-1K in a supervised setting.}
\label{table:jaccard_vit}

\end{table}

\begin{figure}[htpb]
\centering
\includegraphics[width=1\linewidth]{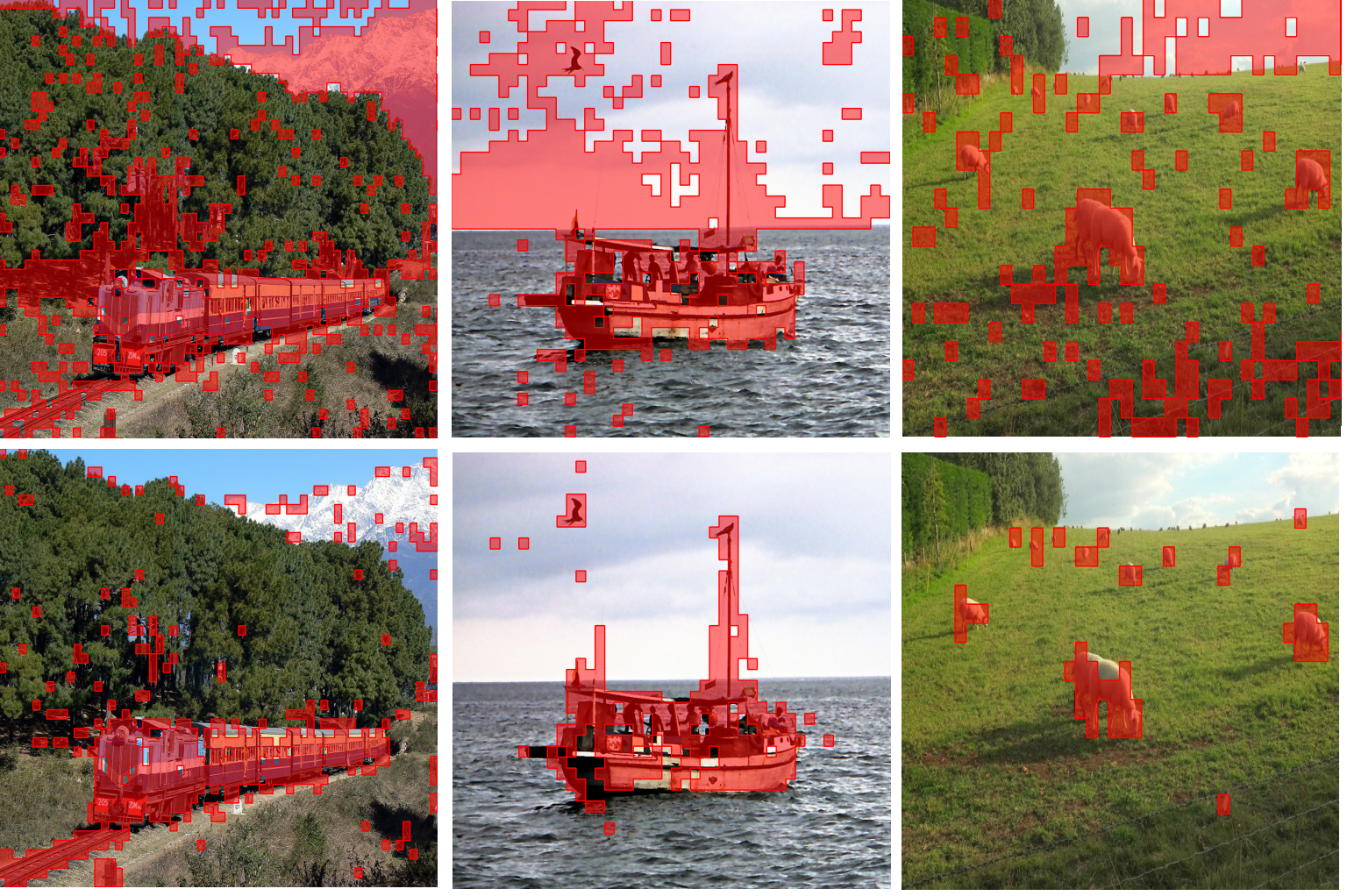}
\caption{\tb{Visualization of segmentation masks}  
using vanilla ViT-S/16 (\emph{top}) and ViT-S +~\methodabbrev~(\emph{bottom}) pretrained supervisedly on ImageNet-1K. We visualize masks obtained by thresholding the self-attention maps to keep $80\%$ of the mass.
}
\label{fig:seg-supervised-pascalvoc}
\vspace*{-6pt}
\end{figure}

\paragraph{Probing self-attention maps in ViTs.} We further analyze whether pretrained ViTs can attend to semantically meaningful regions of the image when evaluated on a different dataset without fine-tuning it. 
We follow the evaluation protocol in~\cite{caron2021emerging}, and visualize the segmentation masks produced from the final layer of the pretrained \methodabbrev~on the Pascal-VOC12~\cite{pascal-voc-2012} validation set. From~\autoref{fig:seg-supervised-pascalvoc}, 
\footnote{The original image sources, before masking, from left to right: \\
\href{https://commons.wikimedia.org/wiki/File:Kangra_Valley_train.jpg}{Kangra valley train} (CC BY-SA 4.0) \\
\href{https://commons.wikimedia.org/wiki/File:Ecuadorian_fishing_boat.jpg}{Ecuadorian fishing boat} (CC BY-SA 2.0) \\
\href{https://www.geograph.org.uk/photo/7354766}{Sheep near Snowshill} (CC BY-SA 2.0)
}
we observe that while vanilla ViT-S/16 does not accurately attend to the object,~\methodabbrev~is able to localize objects quite accurately without any fine-tuning. To quantify this observation,
we follow~\cite{caron2021emerging} and use the Jaccard similarity between predicted segmentation mask and ground truth mask. As shown in~\autoref{table:jaccard_vit}, \methodabbrev~outperforms different variants of vanilla ViT with a significant gap in terms of Jaccard similarity. 
Additionally, we measure the quality of the generated maps for unsupervised object localization using CorLoc~\cite{melas2022deep} as the evaluation metric. From \autoref{table:jaccard_vit}, we observe that \methodabbrev~achieves notable gains across all variants of ViT.

\textbf{Performance on mobile device.}
To verify the efficiency of~\methodabbrev~on low-power devices, we measure its inference time (averaged over 20 iterations) on a Samsung Galaxy S22 device powered by Qualcomm ``Snapdragon\textsuperscript{\textregistered} 8 Gen. 1 Mobile Platform'' with a Qualcomm\textsuperscript{\textregistered} Hexagon\textsuperscript{\tiny{TM}} processor\footnote{Snapdragon and Qualcomm Hexagon are products of Qualcomm Technologies, Inc. and/or its subsidiaries.}, for image resolutions of $224 \times 224$ and $384 \times 384$ using ViT-T/16. 
The inference is performed on Neural Processing Unit in 8-bit precision. As shown in~\autoref{table:snapdragon},~\methodabbrev~improves the runtime by 
$19\%$ for image size of $224\times224$. The gain is even larger at $34\%$ for image resolution  $384 \times 384$, since the number of token increases. 
Thus, skipping computationally-heavy MSA blocks increases throughput by large margins and is confirmed even on mobile hardware.
\begin{table}[h]
\centering
\small
\begin{tabular}{ccc}
\toprule
 \Th{Method} & \Th{$224\times224$} & \Th{$384\times384$} \\
 \midrule
  ViT-T/16       & 5.65 & 20.49 \\ 
  \rowcolor{LightCyan}
  ViT-T/16 + \methodabbrev  & \tb{4.76} & \tb{15.22}\\ 
\bottomrule
\end{tabular}
\vspace{-6pt}
\caption{\textbf{On-device latency} (in msec) of vanilla ViT \vs \methodabbrev~for different image resolutions on a Samsung Galaxy S22 powered by Qualcomm Snapdragon 8 Gen. 1.}
\label{table:snapdragon}
\vspace{-6pt}
\end{table}
\subsection{Self-Supervised Learning with DINO}
\label{sec:exp-dino}
Next, we show the generality of \methodabbrev~as its use in the backbone for self-supervised representation learning (SSL), using DINO~\cite{caron2021emerging}. Since, SSL methods are quite expensive in the pretraining stage in terms of compute and training time, we illustrate that \methodabbrev~achieves comparable performance to using a ViT but with shorter training time. Following the experimental settings of DINO~\cite{caron2021emerging}, we use ViT-S/16~\cite{dosovitskiy2020image} as our student and teacher networks with \methodabbrev~parametric function. We pretrain both baseline and ours using DINO for 100 epochs. We observe that \methodabbrev~achieves almost the same performance as fully trained DINO with around 26\% less training time (73.3\% in 96 GPU-hours \vs 73.6\% in 131 GPU-hours). When trained on 100 epochs, we observe that~\methodabbrev~outperforms DINO by 0.5\% (74.1\% \vs 73.6\%). We show the performance of \methodabbrev~to downstream tasks in the supplementary material.

\subsection{Semantic Segmentation on ADE20K}
\label{sec:exp-segmentation}
\begin{table}
\centering
\scriptsize
\setlength{\tabcolsep}{2pt}
\begin{tabular}{lcccc}
\toprule
 \Th{Method}                                & \Th{Backbone}                             & \Th{mIoU}$\uparrow$     & \Th{GFLOPs}$\downarrow$  & \Th{Throughput}$\uparrow$ \\ 
 \toprule
                                            & ResNet-101~\cite{yu2022metaformer}        & 40.7          & 261         & 24.1 \\
Semantic FPN~\cite{kirillov2019panoptic}    & PoolFormer-S36~\cite{yu2022metaformer}    & 42.0          & 191         & 8.4 \\
                                            & PoolFormer-M36~\cite{yu2022metaformer}    & 42.4          & 271         & 5.4 \\ \midrule
                                            & ResNet-18~\cite{he2016deep}               & 39.9          & 886         & 17.1 \\
                                            & ResNet-101~\cite{he2016deep}              & 44.9          & 1031        & 12.0 \\
                                            & Swin-T~\cite{liu2021swin}                 & 45.8          & 945         & 14.2 \\
                                            & ConvNeXt-T~\cite{liu2022convnet}          & 46.7          & 939         & 15.7 \\
                                            \cmidrule{2-5}
UperNet~\cite{xiao2018unified}              & ViT-T~\cite{dosovitskiy2020image}         & 37.3          & 212          & 24.1 \\
                                            & \CC{100}ViT-T +~\methodabbrev             &\CC{100}\tb{40.6}  &\CC{100}\tb{173}   & \CC{100}\tb{34.7} \\
                                            & ViT-S~\cite{dosovitskiy2020image}         & 44.4          & 360          & 19.5 \\
                                            & \CC{100}ViT-S +~\methodabbrev             & \CC{100}\tb{45.3}  & \CC{100}\tb{283}  & \CC{100}\tb{27.2} \\
                                            & ViT-B~\cite{dosovitskiy2020image}         & 45.6          & 787          & 11.1 \\
                                            & \CC{100}ViT-B +~\methodabbrev             & \CC{100}\tb{46.3}     & \CC{100}\tb{633}  & \CC{100}\tb{15.5} \\
 \toprule
\end{tabular}
\vspace{-6pt}
\caption{\tb{Semantic Segmentation results on ADE20K}. All models are pretrained on ImageNet-1k and fine-tuned on ADE20K. Following Swin~\cite{liu2021swin} and ConvNeXt~\cite{liu2022convnet}, we report mIoU with multi-scale testing. FLOPs and throughput are calculated on the input size of $2048\times512$. Throughput of all models are measured with a batch size of $1$ on a single NVIDIA A100 GPU, averaged over $100$ forward passes.}

\label{table:ade20k_sota}
\vspace{-6pt}
\end{table}
We go beyond classification and show the performance of \methodabbrev~to dense prediction tasks such as semantic segmentation. 
We follow the experimental settings in~\cite{liu2022convnet, liu2021swin} and use MMSegmentation~\cite{mmseg2020} to evaluate \methodabbrev~on ADE20K~\cite{zhou2017scene}.
We observe from~\autoref{table:ade20k_sota}, that \methodabbrev~consistently outperforms all variants of ViT with $15\%$ fewer FLOPs and $25\%$ improved throughput. Interestingly, \methodabbrev-S (ViT-S $+$ \methodabbrev) achieves $8\%$ higher mIoU while being faster than ViT-T. Furthermore, \methodabbrev-S has comparable mIoU with Swin-T~\cite{liu2021swin} whilst having $3\times$ fewer FLOPs and being $1.7\times$ faster. Comparing to fully convolution-based architectures, \methodabbrev-T (ViT-T $+$ \methodabbrev) is on par with ResNet-18 in mIoU while having $4.7\times$ fewer FLOPs and being $1.8\times$ faster.



\subsection{Image Denoising}
\label{sec:exp-denoising}
\methodabbrev~can also generalize to low-level tasks such as image denoising on SIDD~\cite{abdelhamed2018high}, which consists of images with real-world noise. We also demonstrate that \methodabbrev~can generalize to other transformer architectures. In particular, we apply it on Uformer~\cite{wang2022uformer}, a SoTA image denoising model. Uformer is a U-shaped hierarchical network with Swin transformer blocks as the encoder and decoder, and skip connections between them. In~\methodabbrev, we skip window self-attention (WSA) block in each decoder block by reusing attention of the corresponding encoder block via~\methodabbrev~parametric function. Detailed implementation is in the supplementary material. 
%
Following the experimental settings in~\cite{wang2022uformer}, we observe in~\autoref{table:sidd} that \methodabbrev~outperforms the baseline Uformer variants with the 25\% higher throughput on average. Furthermore, we observe that ~\methodabbrev-B (Uformer-B $+$ \methodabbrev) achieves comparable performance with Restormer~\cite{zamir2022restormer}, in terms of PSNR and SSIM, which is the SoTA image denoising method while having $2\times$ fewer FLOPs. Thus, we show the ability of ~\methodabbrev~to generalize to different tasks and also across architectures. 
\begin{table}
\centering
\scriptsize
\begin{tabular}{lcccc}
\toprule
 \Th{Method}                      & \Th{PSNR}$\uparrow$      & \Th{SSIM}$\uparrow$  & \Th{GFLOPs}$\downarrow$   & \Th{Throughput}$\uparrow$ \\ 
 \toprule
 UNet~\cite{ronneberger2015u}         & 39.65      & -          & 35            & --  \\
 DAGL~\cite{mou2021dynamic}           & 38.94      & 0.953      & 255           & --  \\
 DeamNet~\cite{ren2021adaptive}       & 39.47      & 0.957      & 145           & --  \\
 MPRNet~\cite{zamir2021multi}         & 39.71      & 0.958      & 573           & --  \\
 NBNet~\cite{cheng2021nbnet}          & 39.75      & 0.959      & 91            & --  \\
 Restormer~\cite{zamir2022restormer}  & 40.02      & 0.960      & 140           & --  \\ \midrule
 Uformer-T~\cite{wang2022uformer}     & 39.66      & --         & 12            & 17.6   \\
 \rowcolor{LightCyan}
 Uformer-T +~\methodabbrev            & \tb{39.69} & 0.959      & \tb{11}       & \tb{22.2}  \\
 \cmidrule{2-5}
 Uformer-S~\cite{wang2022uformer}     & 39.77      & 0.959      & 44            & 15.1  \\
 \rowcolor{LightCyan}
 Uformer-S +~\methodabbrev            & \tb{39.84} & \tb{0.960} & \tb{39}       & \tb{18.9}  \\
 \cmidrule{2-5}
 Uformer-B~\cite{wang2022uformer}     & 39.89      & \tb{0.960} & 89            & 9.2  \\
 \rowcolor{LightCyan}
 Uformer-B +~\methodabbrev            & \tb{39.94} & \tb{0.960} & \tb{77}       & \tb{10.9}  \\\bottomrule
\end{tabular}
\vspace{-6pt}
\caption{\tb{Image denoising} on SIDD dataset using  PSNR and SSIM~\cite{wang2004image} as the evaluation metrics in the RGB space. FLOPs and throughput are calculated on the input size of $256\times256$, on a single NVIDIA V100 GPU, averaged over the test set of SIDD.}
\label{table:sidd}
\end{table}

\subsection{Video Denoising}
\label{sec:exp-video-denoising}
We further apply our model to the temporal task of video denoising. 
As encoder and decoder backbone, we use UniFormer~\cite{li2022uniformer}, a U-shaped hybrid encoder-decoder architecture with 3D convolutions and spatio-temporal global self-attention blocks. Detailed implementation is provided in the supplementary material.
Similar to image denoising, we skip MSA blocks in the decoder, however, simply adopt a naive \methodabbrev, where we reuse window self-attention matrix, $A$, of the corresponding encoder block using an Identity function. We empirically observe that reusing attention  works better in this task, and shows the ability of our method to be applied for different scenarios.  
We follow the experimental settings in ~\cite{tassano2020fastdvdnet} and train \methodabbrev~on DAVIS~\cite{Pont_Tuset_arXiv_2017} dataset. We train using Charbonnier loss~\cite{charbonnier1994two} on patches of $7\times128\times128$ using a multiple-input, multiple-output (MIMO) paradigm (i.e. the model outputs $7$ reconstructed frames from $7$ input frames) for noise level $\sigma = 30$. 
From~\autoref{table:davis}, we observe that \methodabbrev~ performs on par with baseline Uniformer, while having 17\% fewer FLOPs. This shows that \methodabbrev~can generalize to temporal tasks.
\begin{table}
\centering
\scriptsize
\setlength{\tabcolsep}{4pt}
\begin{tabular}{l|ccccc}
\toprule
 \Th{Method}  & FastDVDNet                   & PaCNet                  & VRT                 & UniFormer               & UniFormer+    \\
              & \cite{tassano2020fastdvdnet} & \cite{vaksman2021patch} & \cite{liang2022vrt} &  \cite{li2022uniformer} & \methodabbrev \\
 \midrule
 \Th{PSNR}$\uparrow$    & 34.04                        & 34.79                   & 36.52               & 35.24                   & 35.16 \\
 \Th{GFLOPS}$\downarrow$  & 41.9                         & 34.8                    & 708.8               & 93.2                    & 77.1 \\
\bottomrule
\end{tabular}
\vspace{-6pt}
\caption{\tb{Video denoising} Quantitative comparison (average RGB channel PSNR) with state-of-the-art methods for video denoising on DAVIS, with additive noise level $\sigma=30$. FLOPs are calculated per frame per patch size of $256\times256$.}
\label{table:davis}
\vspace{-6pt}
\end{table}

\begin{table}
\centering
\scriptsize
\setlength{\tabcolsep}{4pt}
\begin{tabular}{ccccc}
\toprule
 \Th{Function}        & \Th{kernel}   & \Th{Channel}      & \Th{Top-1}$\uparrow$ & \Th{throughput}$\uparrow$ \\ 
  $\Phi$              &               & \Th{expansion}    &   ($\%$)   & (img/sec $\times 10^3$)\\
\midrule
ViT-T               & -             & -              & 65.8       & 5.8   \\
\midrule 
 \textsc{Identity}  & -             & -              & 61.1       & 8.5 \\
 \textsc{Conv}      & $5\times5$    & -              & 65.4       & 5.2 \\
 \textsc{DwC}       & $5\times5$    & -              & 65.6       & 7.8 \\
 \midrule
                    & $3\times3$    &                & 67.1       & 7.3 \\
 \methodabbrev      & $5\times5$    & $2$            & \tb{67.7}  & 6.9 \\
                    & $7\times7$    &                & 67.4       & 6.6 \\
 \midrule
                    &               & $0.5$          & 64.4       & 7.4 \\
 \methodabbrev      & $5\times5$    & $1$            & 65.9       & 7.2 \\
                    &               & $2$            & \tb{67.7} & 6.9 \\
 \bottomrule
\end{tabular}
\vspace{-6pt}
\caption{\tb{Ablations} using ViT-T/16 on ImageNet-1K for 100 epochs. We measure throughput (image/sec) with a batch size
of 1024 on a single NVIDIA A100 GPU, averaged over the validation set of ImageNet-1K. }
\label{table:ablation}
\vspace{-6pt}
\end{table}

\subsection{Ablations}
\label{sec:ablations}
All ablations are performed using ViT-T/16 on ImageNet-1K for 100 epochs to reduce the training time. Unless specified, following \methodabbrev~we skip the MSA blocks from layer 3 through 8 for all ablations. 

\textbf{Parametric function $\Phi$. } We study the effect of different parametric functions in terms of accuracy and throughput. As discussed in~\autoref{sec:param-prop}, $\Phi$ can be as simple as an identity function, where we directly reuse representations from a previous MSA block into one of more subsequent MSA blocks. 
From~\autoref{table:ablation}, using an identity function results in a 4.7\% drop in top-1 accuracy while being 47\% faster than baseline ViT. Using a convolution or DwC~\cite{chollet2017xception} with kernel size $5 \times 5$ as a parametric function leads to the same performance as the baseline. However, DwC is 0.2\% better and 50\% faster than convolution, and 34\% faster than the baseline.~\methodabbrev~parametric function outperforms all.

\textbf{Kernel size. } By default~\methodabbrev~uses a DwC with kernel size of $5 \times 5$. 
As shown in~\autoref{table:ablation}, while using a $3 \times 3$ kernel is faster than default \methodabbrev~by 6\%, it is 0.6\% worse in terms of accuracy. A larger kernel size has poor accuracy and lower throughout. However, irrespective of the kernel size, \methodabbrev~outperforms the baseline ViT-T by at least 1.4\%, showing its ability to encode cross-token interactions.

\textbf{Channel expansion. } In the \methodabbrev~, the first linear layer $\text{FC}_1$, expands the channel dimension from $d \to 2d$. 
\autoref{table:ablation} shows the impact of channel dimension, \ie, when the  channel expansion ratio of $\text{FC}_1$ is $1.0$ ($d \to d$) and 0.5 ($d \to d/2$). We observe that while the lower channel expansion ratio improves the throughput, it performs worse than default~\methodabbrev. This could be due to sub-optimal representations encoded by the DwC due to fewer filters.  

\textbf{Skipping MSA in alternate configuration. } Instead of skipping the MSA operation in the layers  $3 - 8$, we study the effect of skipping MSA operation at $l \in \{3,5,7,9\}$.
We observe the latter configuration outperforms the baseline ViT by 2.7\% (65.8 \vs 67.5\%). However, it performs 0.2\% lower and is 8\% slower than our default~\methodabbrev~configuration. 

\section{Conclusion}
We proposed~\methodabbrev, a plug-in module that can be placed in any ViT architecture for reducing the costly 
Self-Attention computations.~\methodabbrev~leverages the dependency across MSA blocks and bypasses attention computation by re-using attention from previous MSA blocks. To ensure that the metaphorical sharing is caring we introduced a simple and light parametric function that does not affect the inductive bias encoded in MSA. The~\methodabbrev~function is able capture cross-token relations and outperforms the baseline while being computationally faster in terms of throughput and FLOPs. We plugged~\methodabbrev~into different transformer architectures and showed its effectiveness on 7 different tasks.
\label{sec:conclusion}

{\small
\bibliographystyle{ieee_fullname}
\bibliography{bib}
}

\clearpage



\appendix

\section{Implementation details}
\label{sec:exp-settings}


\subsection{Hyper-parameters}
\paragraph{ImageNet-1K: Image classification.}
We train \methodabbrev~on the ILSVRC-2012  dataset~\cite{deng2009imagenet} with 1000 classes (referred as ImageNet-1K). We follow the experimental settings of DeIT~\cite{touvron2021training} and use the codebase from the timm library~\cite{rw2019timm} to train ViT-T, ViT-S and ViT-B. We use the default $16 \times 16$ patch size, using an image resolution of $224 \times 224$ with total number of tokens $n = 196$. We train baseline ViT and \methodabbrev~for 300 epochs from scratch on 4 NVIDIA A100 GPUs using batch sizes of 2048 for ViT-T and 1024 for ViT-S and ViT-B. 

\paragraph{ImageNet-1K: Self-supervised learning.}
We follow the experimental settings of DINO~\cite{caron2021emerging} and pre-train DINO and \methodabbrev~on ImageNet-1K using ViT-S/16 as the backbone. While likely the hyperparameters could be tuned further for our proposed \methodabbrev~ method, we use same hyper-parameters for both the baseline and ours, yielding a conservative estimate of our model's performance. 
We pre-train both methods from scratch for 100 epochs using 4 NVIDIA A100 GPUs. For linear-probing, we freeze the backbone from the pre-training stage and fine-tune the classifier for 100 epochs, exactly as done in~\cite{caron2021emerging}.

\paragraph{Pascal-VOC2012: Unsupervised object segmentation.}
We use the Pascal VOC 2012~\cite{pascal-voc-2012} validation set for this experiment, containing $1449$ images. We follow DINO and obtain unsupervised segmentation masks by thresholding the averaged self-attention map (extracted from the last layer of a pretrained ViT/\methodabbrev~model) to keep $80\%$ of the mass.
The Jaccard similarity $J$ between a predicted mask, $P$, and ground-truth mask, $G$, is defined as:
$$ J (P, G) = \frac{G \cap P}{G \cup P} $$
We report Jaccard similarity, averaged over all the samples.

\paragraph{ADE20K: Semantic segmentation.}
We evaluate~\methodabbrev~on ADE20K~\cite{zhou2017scene}, a widely-used semantic segmentation dataset, covering $150$ semantic categories. The dataset includes 20K and 2K images in the training and validation set, respectively. Different variants of~\methodabbrev~are evaluated using UperNet~\cite{xiao2018unified} as the backbone. 
We use our ImageNet-1K pretrained model to initialize the backbone and Kaiming~\cite{he2015delving} initialization for other layers. We use AdamW~\cite{loshchilov2017decoupled}, with an initial learning rate of $6e-5$, weight decay of $1e-2$, and linear warmup of $1500$ iterations. All models are trained for $160K$ iterations with a batch size of $16$ using MMSegmentation repo~\cite{mmseg2020}. We keep the same hyper-parameters for~\methodabbrev~and ViT. 

\paragraph{SIDD: Image denoising.}
We follow the experimental settings in Uformer~\cite{wang2022uformer} and train \methodabbrev~on the Smartphone Image Denoising Dataset (SIDD)~\cite{SIDD_2018_CVPR} which consists of real-world noise. The training samples are first randomly cropped to $128 \times 128$ patches and input to the model, which is trained for 250 epochs using batch size 32. The model is then evaluated on images of size $256 \times 256$.

\paragraph{DAVIS: Video denoising.}
We further apply our model to the temporal task of video denoising. 
We adopt the same U-shape encoder-decoder based architecture of UFormer. As the encoder and decoder backbone, we use UniFormer~\cite{li2022uniformer}.
We train the model on noise level $\sigma = 30$ using Charbonnier loss~\cite{charbonnier1994two} on patches of $7\times128\times128$ using a multiple-input, multiple-output (MIMO) paradigm~\cite{liang2022vrt} (\ie, the model outputs $7$ reconstructed frames from $7$ input frames). During inference, a video is divided into 3D patches of $7\times128\times128$ with an overlap of $10$ pixels. Each patch is fed to the model and the outputs are merged to obtain the final denoised video. Following~\cite{tassano2020fastdvdnet}, PSNR is calculated as averaged over videos.
We use the same training hyper-parameters as image denoising. 

\subsection{Architecture}
\label{sec:supp-arch}
\paragraph{Image Classification.}
All baseline ViT variants have 12 layers in total, which remains unchanged with \methodabbrev. Following the CKA analysis of $Z^{\text{MSA}}$ in Figure 3(b) of our main paper, we skip computing the MSA blocks in layer 3 through 8 for all ViT variants and retrain it from scratch.

\paragraph{Image Denoising.}
We apply \methodabbrev~to Uformer~\cite{li2021uniformer} a SoTA image denoising model. Uformer is a U-shaped hierarchical network with Swin transformer blocks as the encoder and decoder, and skip connections between them. In \methodabbrev, we skip window self-attention (WSA) block in each decoder block by reusing attention of the corresponding encoder block via \methodabbrev~parametric function. Let $Z_{l}^{\text{WSA}_e} \in \real^{n \times c}$ denote the output of the WSA block at layer $l$ from the encoder and $Z_{l-1}^d \in \real^{n \times c}$ denote the output of the layer $l-1$ from the decoder of Uformer. The input to the WSA block (which is skipped) at layer $l$ of the decoder is given by
\begin{align}
\hat{Z}_{l}^{\text{WSA}_d} = \Phi(Z_{l}^{\text{WSA}_e}; Z_{l-1}^d)  \in \real^{n \times 2c}  
\end{align}

\begin{figure}
\centering
\includegraphics[width=1\linewidth]{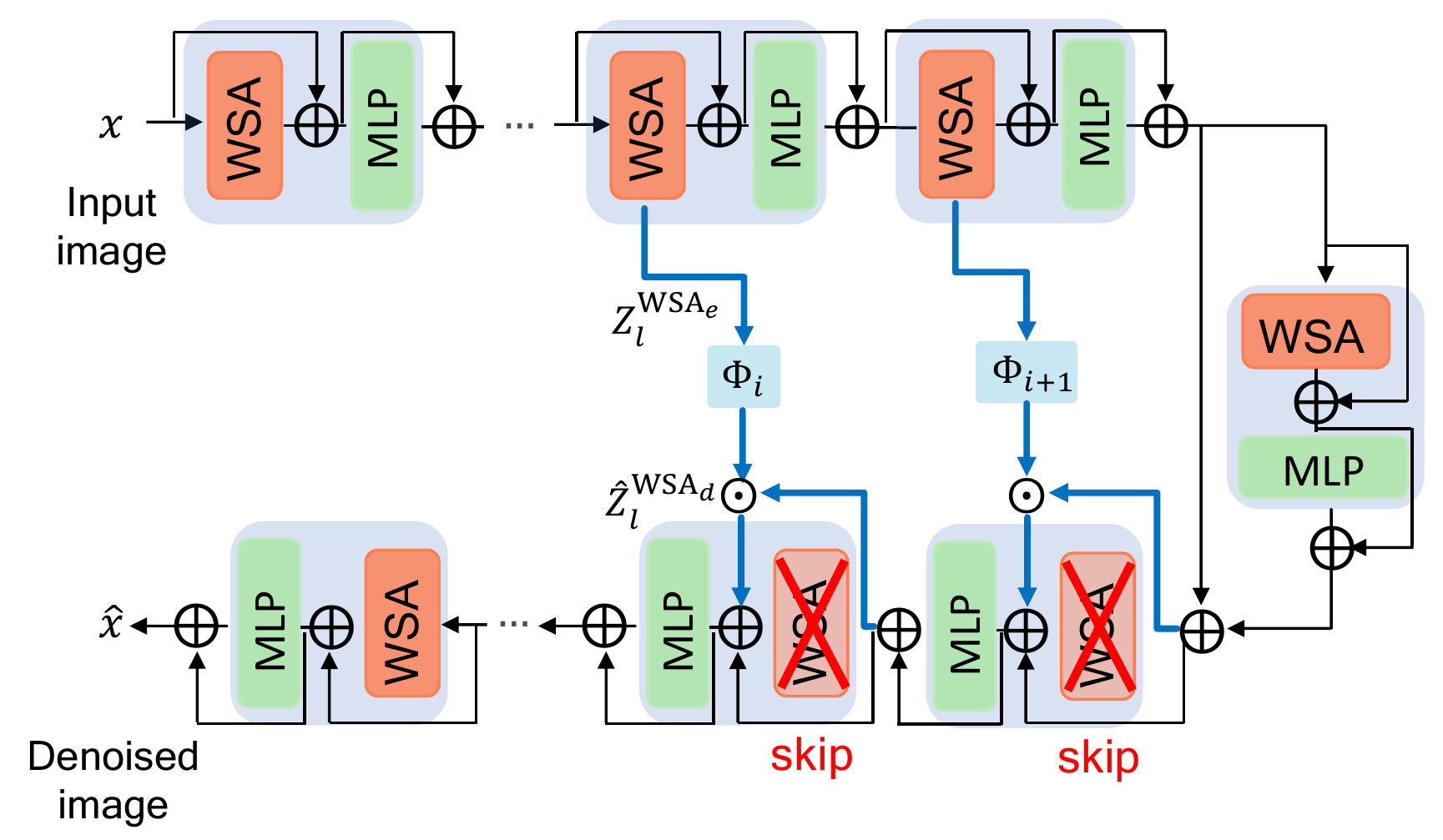}
\caption{\tb{Framework of \methodabbrev~on Uformer} Instead of standard MSA block in ViT, Uformer uses window self-attention (WSA) block similar to Swin Transformer. We skip WSA block in the layers close to the bottleneck. 
}
\label{fig:uformer-skat}
\vspace{-6pt}
\end{figure}

Here, ``;" denotes concatenation along the channel dimension. We show the framework of \methodabbrev~on Uformer in \autoref{fig:uformer-skat}

\paragraph{Video Denoising.}
we apply \methodabbrev~to UniFormer~\cite{li2022uniformer}, a U-shaped hybrid encoder-decoder architecture with 3D convolutions and spatio-temporal global self-attention blocks. The encoder of UniFormer comprises two 3D convolution layers followed by two spatio-temporal transformer layers with global self-attention (MSA) blocks. A downsampling operation is used after every layer in the encoder. The decoder is symmetric to the encoder with two transformer layers followed by two 3D convolution layers with an upsampling operation between each layer. Similar to Uformer, skip connections are used between encoder and decoder. Similar to image denoising, we skip MSA blocks in the decoder, however, simply adopt a naive SKAT, where we reuse global self-attention matrix $A_l$, from the encoder at layer $l$ in the corresponding decoder stage at the same layer using an Identity function.  Let $A_{l}^e \in \real^{n \times n}$ denote the self-attention matrix at layer $l$ from the encoder. The self-attention in the decoder stage at layer $l$ is given by $A_{l}^d = I(A_{l}^e) \in \real^{n \times n}$, where $I(.)$ is the identity function.

We observe that skipping attention $A$ using an identity function works better than skipping MSA blocks using the \methodabbrev~parametric function. This shows the generality of \methodabbrev, regardless of approximating attention or MSA blocks.


\section{Additional experiments}
\paragraph{Image classification.}
\begin{table}
\centering
\scriptsize
\setlength{\tabcolsep}{3pt}
\begin{tabular}{llcccc} \toprule
 \Th{Backbone}    & \Th{Method}                             &\Th{top-1}   & \Th{Param}        & \Th{GFlops}   & \Th{Throughput}          \\ 
                  &                                         & (\%)        & ($\times 10^6$)   &               & (img/sec $\times 10^3$)  \\ \midrule
                  \cmidrule{2-6}
                  & T2T-ViT~\cite{yuan2021tokens}           & 71.7        & 5.8               & 1.1           & --                         \\
                  & ConvNeXt (iso)~\cite{liu2022convnet}    & 72.7        & 5.7               & 1.1           & 5.8                      \\
                  \cmidrule{2-6}
                  & ViT~\cite{dosovitskiy2020image}         & 72.8        & 5.7               & 1.2           & 5.8                      \\
                  & A-ViT~\cite{yin2022vit}                 & 71.0        & 5.7               & 0.8           & 6.3                      \\
                  & Dynamic ViT~\cite{rao2021dynamicvit}    & 70.9        & --                & 0.9           & 6.1                      \\
ViT-T/16          & SViTE~\cite{chen2021chasing}            & 71.7        & \tb{4.0}          & 0.9           & 6.2                      \\
                  & SPViT~\cite{kong2021spvit}              & 72.7        & 5.7               & 0.9           & 6.7                      \\
                  & ATS~\cite{fayyazadaptive}               & 72.7        & 5.7               & 0.9           & 6.1                      \\
                  & PS-ViT~\cite{tang2022patch}             & 72.6        & --                & \tb{0.7}      & 6.6                      \\
                  & HVT~\cite{pan2021scalable}              & 70.2        & 5.7               & \tb{0.7}      & \tb{7.2}                 \\
                  \cmidrule{2-6}
                  \rowcolor{LightCyan}
                  & ~\methodabbrev                          & \tb{72.9}   & 5.8               & 1.1           & 6.9                      \\
                  \midrule 
                  & ConvNext-T~\cite{liu2022convnet}       & \tb{82.1}   & 29.0              & 4.5           & 2.6                      \\
                  & ConvNeXt (iso)~\cite{liu2022convnet}   & 79.7        & 22.4              & 4.3           & 3.3                      \\
                  & Swin-T~\cite{liu2021swin}              & 81.3        & 28.3              & 4.5           & 2.5                      \\
                  & T2T-ViT~\cite{yuan2021tokens}          & 80.7        & 21.5              & 5.2           & --                      \\
                  & CoaT-Lite-Small                        & 81.9        & 20.0              & 4.0           & --                         \\
                  & Poolformer-S24~\cite{yu2022metaformer} & 80.3        & 21.0              & 3.4           & --                         \\
                  & Twins-SVT-S~\cite{chu2021twins}        & 81.7        & 24.0              & 2.8           & --                        \\
                  & MobileViT-S~\cite{mehta2021mobilevit}  & 78.4        & \tb{5.6}          & \tb{2.0}      & --                           \\
                  & PVT~\cite{wang2021pyramid}              & 79.8        & 24.5              & 3.8           & --                       \\
                  \cmidrule{2-6}
ViT-S/16          & ViT~\cite{dosovitskiy2020image}         & 79.8        & 22.4              & 4.6           & 3.2                      \\
                  & A-ViT~\cite{yin2022vit}                 & 78.6        & 22.4              & 3.6           & 3.4                      \\
                  & Dynamic ViT~\cite{rao2021dynamicvit}    & 78.3        & 23.1              & 3.4           & 3.6                      \\
                  & SViTE~\cite{chen2021chasing}            & 80.2        & 13.1              & 2.7           & 3.5                      \\
                  & ATS~\cite{fayyazadaptive}               & 79.7        & 22.4              & 2.9           & 3.3                      \\
                  & PS-ViT~\cite{tang2022patch}             & 79.4        & --                & 2.6           & 3.9                      \\
                  & SPViT~\cite{kong2021spvit}              & 79.3        & 22.1              & 2.7           & 3.5                      \\
                  & Rev-ViT~\cite{mangalam2022reversible}   & 79.8        & 22.4              & 4.6           & 3.6                      \\
                  & HVT\cite{pan2021scalable}               & 78.0        & 22.5              & 2.4           & \tb{4.1}                  \\
                  \cmidrule{2-6}
                  \rowcolor{LightCyan}
                  &  ~\methodabbrev                         & 80.2        & 22.1              & 4.0           & 3.8                      \\
                  \midrule
                  \midrule
                   & Swin-S~\cite{liu2021swin}              & \tb{83.5}   & 88.0              & 15.4          & 1.0                      \\
                   & Twins-SVT-B~\cite{chu2021twins}        & 83.2        & 56.0              & 8.6           & --                        \\
                   & PVT~\cite{wang2021pyramid}             & 81.7        & 61.4              & \tb{9.8}      & --                         \\
                  & ConvNeXt (iso)~\cite{liu2022convnet}    & 82.0        & 87.3              & 16.9          & 1.3                      \\
                  \cmidrule{2-6}
ViT-B/16          & ViT~\cite{dosovitskiy2020image}         & 81.8        & 87.3              & 17.6          & 1.2                      \\
                  & SViTE~\cite{chen2021chasing}            & 81.6        & \tb{52.0}         & 11.5          & 1.3                         \\
                  & Rev-ViT~\cite{mangalam2022reversible}   & 81.5        & 87.3              & 17.6          & 1.2                         \\
                  & PS-ViT~\cite{tang2022patch}             & 81.5        &  --               & \tb{9.8}      & \tb{1.6}                         \\
                  \cmidrule{2-6}
                  \rowcolor{LightCyan}
                  &~\methodabbrev                           & 82.2        & 86.7              & 15.2          & 1.5                       \\
                  \bottomrule
\end{tabular}
\vspace{-6pt}
\caption{\tb{Image classification on ImageNet-1K.} Accuracy \vs efficiency comparison of \methodabbrev~with SoTA methods for image resolution $224 \times 224$. For all the methods, we measure throughput (image/sec) with a batch size of 1024 on a single NVIDIA A100 GPU, averaged over the validation set of ImageNet-1K.}
\label{table:sota-imagenet-supp}
\end{table}
Here we extend our SoTA comparison with methods that go beyond vanilla ViT architectures. These methods include hierarchical (Swin, PVT, Poolformer, MobileViT, Twins-SVT) and Hybrid (ConvNext, CoAT) architectures.
We provide the complete set of SoTA methods that improve the efficiency of ViT either by token sampling (extending Table 1 in our main paper), using hybrid architectures or window self-attention blocks in \autoref{table:sota-imagenet-supp}. Apart from methods that perform efficient token sampling, none of the other methods are directly comparable because they modify the underlying architecture of ViT, either by using window self-attention blocks or reducing the overall number of transformer layers. 

\paragraph{Unsupervised segmentation of DINO.}
We follow DINO~\cite{caron2021emerging} and evaluate the performance of baseline DINO \vs \methodabbrev~on unsupervised object segmentation on Pascal-VOC2012~\cite{pascal-voc-2012} dataset. We follow the experimental setting as discussed in \autoref{sec:exp-settings} and observe that baseline DINO has a Jaccard similarity of $45.3$ while \methodabbrev~achieves $44.7$. While \methodabbrev~outperforms DINO on image classification by $0.5$\%, we achieve comparable performance in terms of unsupervised object segmentation.


\section{Additional ablations}

\paragraph{Reusing self-attention.}
As mentioned in Subsection 3.3, we skip the $Z^{\text{MSA}}$ in \methodabbrev~as the compute and memory benefit from skipping the entire MSA block is greater than skipping just the self-attention operation. Here we study the effect of skipping just the self-attention operation.  Let $A_{l-1}$ denote the self-attention matrix at layer $l-1$, then the self-attention matrix at layer $l$ is given by $\hat{A}_l = I(A_{l-1})$. Similar to \methodabbrev~we skip computing the self-attention matrix from layers 3 through 8. As parametric function $\Phi$, we use an identity mapping and train ViT-T/16 from scratch for 100 epochs on ImageNet-1K. We observe from \autoref{tab:supp-ablation}, that skipping the self-attention matrix results in a top-1 accuracy of $63.2$\% which is $2.1$\% higher than the skipping $Z^{\text{MSA}}$ with an identity function (61.1\% - Table 7 of main paper). However, skipping self-attention matrix results in $20$\% decrease in throughput (8500 $\rightarrow$ 6800 images/sec) as compared to using an identity function to skip MSA block.
It is interesting to note that 
skipping self-attention matrix results in a lower drop in performance as compared to skipping MSA block.
However, applying a parametric function to skip self-attention can be challenging due to the properties of the self-attention matrix, and we leave this to future work. 

\paragraph{\methodabbrev~in pretrained model.}
As mentioned in \autoref{sec:supp-arch}, we train \methodabbrev~with all variants of ViT from scratch. For completeness, we also study the effect of skipping the self-attention matrix and the MSA block on a pretrained ViT-T using an Identity function, \textit{without retraining}. We observe from \autoref{tab:supp-ablation} that skipping the self-attention computation in layers 3 through 8, results in a top-1 accuracy of $53.9$\%, while skipping MSA blocks results in top-1 accuracy of $47.8$\%. It is interesting to note that the drop in top-1 accuracy from skipping self-attention is merely 19\%  (72.8 $\rightarrow$ 53.9) on average and does not result in an \textit{extremely} large drop as one might expect. This shows that there indeed exists high correlation across self-attention and $Z^{\text{MSA}}$, which \methodabbrev~utilizes to improve the efficiency of the ViTs.

\begin{table}[h]
\centering
\small
\begin{tabular}{l|c|cc}
\toprule
 \Th{Method} & \Th{Training} & \Th{top-1 (\%)} & \Th{throughput}\\
 \midrule
  $A$                   & \cmark & 63.2 & 6800 \\
  $Z^{\text{MSA}}$      & \cmark & 61.1 & 8500 \\
  \midrule
  $A$                   & \xmark & 53.9 & 6800 \\
  $Z^{\text{MSA}}$      & \xmark & 47.8 & 8500 \\
\bottomrule
\end{tabular}
\vspace{-6pt}
\caption{\textbf{Ablations} on the effect of skipping the self-attention, $A$, and the MSA block, $Z^{\text{MSA}}$. In the first two rows, models are trained for 100 epochs. In the last two rows we use a pretrained ViT-T/16 and simply skip computations in blocks 3-8 during inference. For all the experiments with use Identity function as $\Phi$.}
\label{tab:supp-ablation}
\vspace{-6pt}
\end{table}

\section{CKA analysis of attention from ViT-T}

\begin{figure}
\centering
\includegraphics[width=.8\linewidth]{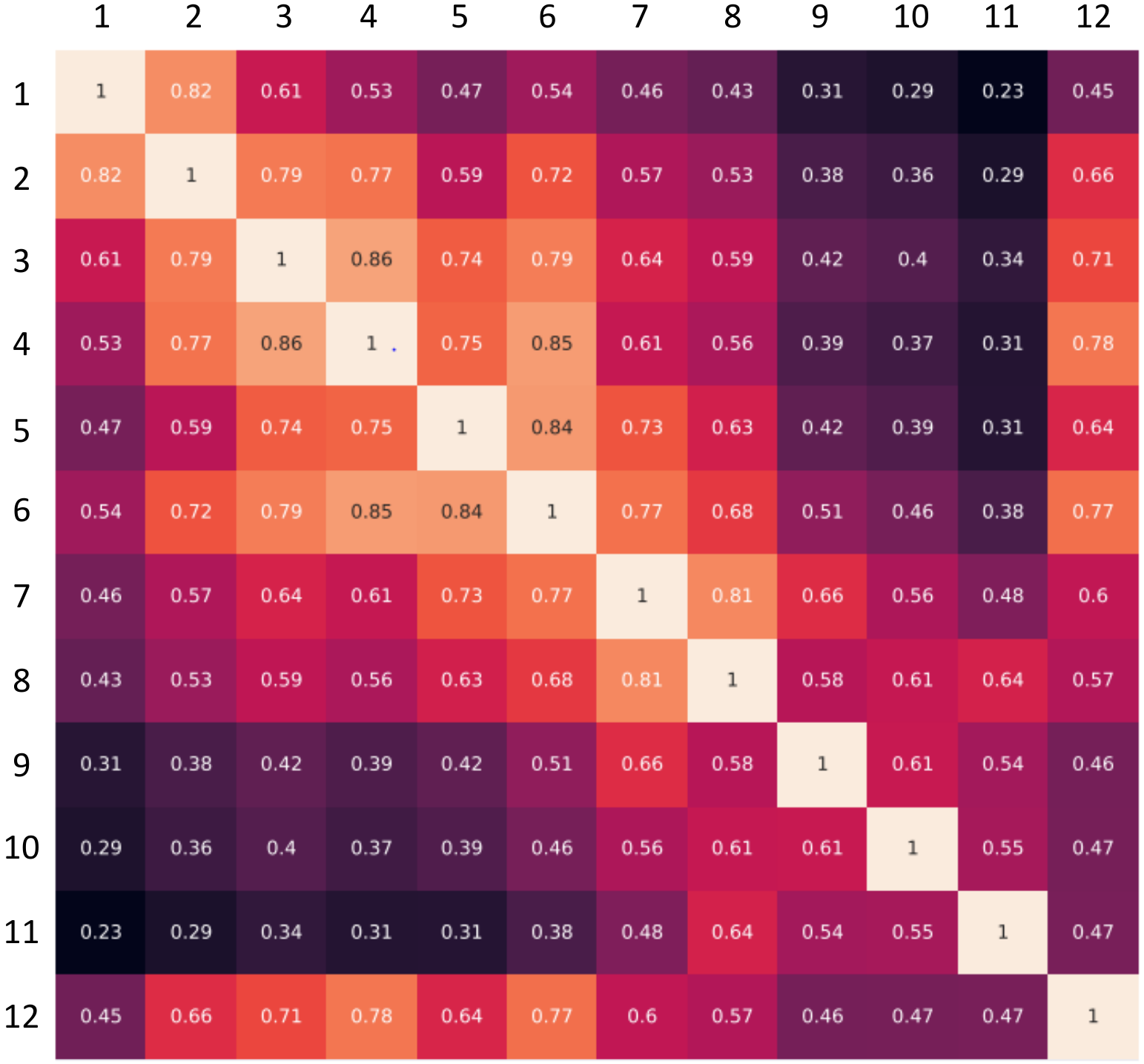}
\caption{\tb{CKA analysis of $A$} for all tokens from pretrained vanilla ViT-T/16 on the validation set of ImageNet-1K. We observe a  high correlation for all tokens in $A$ from layers 1 to 8.
}
\label{fig:cka-alltokens}
\vspace{-6pt}
\end{figure}

As discussed in Section 3.2 of our main paper, we analyze the CKA of the self-attention matrix for all tokens between different layers of ViT-T/16 pretrained on ImageNet-1K. Since in the supervised setting $A \in \real^{(n+1) \times (n+1)}$, we first remove the \texttt{CLS} token to obtain $A^P \in \real^{n \times n}$. We then compute the CKA of $A^{P}_{l}$ for $l \in \cL$. We observe from~\autoref{fig:cka-alltokens}, that there exists a high correlation across all the tokens from the self-attention matrix. Thus, reusing self-attention from different layers of the ViT can improve the overall throughput while yielding comparable accuracy as the baseline ViT.

\end{document}